\documentclass[11pt]{article}

\usepackage[preprint]{acl}

\makeatletter
\g@addto@macro{\UrlBreaks}{%
  \do\a\do\b\do\c\do\d\do\e\do\f\do\g\do\h\do\i\do\j\do\k\do\l\do\m
  \do\n\do\o\do\p\do\q\do\r\do\s\do\t\do\u\do\v\do\w\do\x\do\y\do\z
  \do\A\do\B\do\C\do\D\do\E\do\F\do\G\do\H\do\I\do\J\do\K\do\L\do\M
  \do\N\do\O\do\P\do\Q\do\R\do\S\do\T\do\U\do\V\do\W\do\X\do\Y\do\Z
  \do\0\do\1\do\2\do\3\do\4\do\5\do\6\do\7\do\8\do\9}
\mathchardef\UrlBreakPenalty=100
\mathchardef\UrlBigBreakPenalty=100
\makeatother

\usepackage{times}
\usepackage{latexsym}
\usepackage{amsmath}
\usepackage{amssymb}
\usepackage{amsthm}
\newtheorem{proposition}{Proposition}

\usepackage[T1]{fontenc}

\usepackage[utf8]{inputenc}

\usepackage{microtype}

\usepackage{inconsolata}

\usepackage{graphicx}
\usepackage{booktabs}
\usepackage{colortbl}
\usepackage{tabularx}
\usepackage{xspace}
\definecolor{bandblue}{RGB}{230,247,254}
\definecolor{rowgray}{RGB}{242,242,242}
\usepackage{algorithm}
\usepackage{algorithmic}

\newcommand{\lisao}{LiSAO\xspace}
\newcommand{\liisa}{Li\textsuperscript{2}SAO\xspace}

\title{\texorpdfstring{\liisa}{Li2SAO}: Lifting Light Single-Rollout Asynchronous RL Optimization via Group-Relative Actor-Critic Bootstrapping}

\makeatletter
\newcommand{\authormark}[1]{\textsuperscript{\ensuremath{#1}}}
\newcommand{\authornote}[2]{%
  \g@addto@macro\@thanks{%
    \begingroup
      \def\@thefnmark{\ensuremath{#1}}%
      \@footnotetext{#2}%
    \endgroup}}
\makeatother

\author{%
  \textbf{Zichen Bai\authormark{\diamond},
  Yanli Wang\authormark{\diamond},
  Yang Liu\authormark{\diamond},
  Junhao Shi,
  Xueqin Liu,
  Jinguo Sun}\\
  \textbf{Muye Li,
  Shaoqiang Zheng\authormark{\dagger},
  Xuefei Diao,
  Zhuwei Zeng,
  Nan Duan,
  Peng Cao}\\[0.35em]
  JD.Com%
  \authornote{\diamond}{JD Tech Genius Team}%
  \authornote{\dagger}{Corresponding author}%
}

\begin{document}
\maketitle
\begin{abstract}
Asynchronous reinforcement learning (RL) is becoming a mainstream paradigm for post-training large language models (LLMs), as it improves training efficiency by updating the model as rollouts arrive.
Recent work argues that group-wise sampling in GRPO does not naturally fit asynchronous RL training because rollout lengths follow a long-tailed distribution, and proposes Single-rollout Asynchronous Optimization (SAO) to improve stability and off-policy behavior, especially on agentic tasks.
However, SAO relies on a closed-source value model obtained from scaling value pretraining.
In this paper, we analyze GRPO and \lisao, our lightweight reproduction of SAO without the pretrained value model, on single-turn reasoning and multi-turn agentic tasks.
We find that GRPO outperforms \lisao on reasoning tasks, while the reverse holds on agentic tasks.
Since the value model of \lisao is not yet well trained , \lisao lags behind GRPO substantially in early training steps.
To address this issue, we propose \textbf{\liisa}, which \textbf{Li}fts \textbf{Li}ght \textbf{S}ingle-rollout \textbf{A}synchronous \textbf{O}ptimization with group-relative actor-critic bootstrapping: while the critic is cold, a group-relative advantage trains the actor and a group-centered return trains the critic, after which training switches to single-rollout GAE.
Experiments show that \liisa catches up with or surpasses GRPO in early training and further raises the later performance ceiling.
Code is available at \url{https://github.com/jd-cc-coder/Li2SAO}.
\end{abstract}

\section{Introduction}
\label{sec:intro}
Reinforcement learning (RL) has become a central approach for eliciting long chain of thought reasoning and multi-turn agentic behavior in large language models (LLMs), including tool use and environment interaction
\citep{ouyang-2022-instructgpt,shao-2024-deepseekmath,deepseekai-2025-r1}.
Most existing RL systems couple generation and training by completing a full batch of trajectories under a fixed policy before each update, so variable lengths on long reasoning and tool use tasks force fast rollouts to wait for stragglers
\citep{yao-2023-deepspeedchat,hu-2024-openrlhf,sheng-2025-hybridflow}.
Asynchronous RL removes this coupling by running rollout generation and training concurrently, allowing trajectories to enter training as they become available without draining the rollout workers, which substantially improves training efficiency
\citep{noukhovitch-2025-asynchronous,sheng-2025-hybridflow,zhong-2025-streamrl,wu-2025-llamarl}.
A fully asynchronous extension is the \emph{in-flight} weight-update scheme, in which rollout workers receive updated parameters and continue unfinished trajectories without discarding generated tokens
\citep{fu-2025-areal,piche-2025-pipelinerl}.

However, asynchronous RL introduces two challenges.
First, under the \emph{in-flight} weight-update scheme, a trajectory can span multiple policy versions, compounding policy lag with within trajectory distribution shift and destabilizing optimization
\citep{huang2026stableasync}.
Second, the widely used Group Relative Policy Optimization (GRPO) is structurally mismatched to asynchronous training
\citep{shao-2024-deepseekmath,deepseekai-2025-r1}.
GRPO must wait for all responses to the same prompt before computing group relative advantages, which recreates a synchronization barrier and increases the staleness of early responses.
Its sequence level advantage also provides limited temporal credit assignment for long agentic trajectories with multiple decisions and environment interactions.

To address these challenges, \citet{hou-2026-sao} proposes Single-rollout Asynchronous Optimization (SAO), which replaces group wise sampling with one rollout per prompt and consumes each trajectory as soon as it arrives.
SAO stabilizes off policy updates with Direct Double Sided Importance Sampling (DIS) and supports agentic credit assignment with Skip Observation token level GAE.
However, controlling the high variance of single rollout optimization requires a well calibrated value model, and SAO relies on scaled value pretraining whose model and construction pipeline are not publicly available.
This raises a natural question:
\begin{center}
\emph{\textbf{How can we recover the benefits of SAO \newline without a pretrained value model?}}
\end{center}

To answer this question, we first reproduce a lightweight form of SAO (\lisao) without value pretraining.
Our implementation is built on the open source \textsc{slime} RL framework\footnote{\url{https://github.com/THUDM/slime}} and we implement the \emph{in-flight} weight-update asynchronous training for both single-turn reasoning and multi-turn sandbox-based agentic tasks.
We find that GRPO and \lisao exhibit complementary strengths, with GRPO performing better on single-turn reasoning and \lisao on multi-turn agentic tool use.
More importantly, \lisao consistently lags behind GRPO during early optimization in both settings.
Our analysis attributes this early gap to critic calibration.
A critic initialized from the policy checkpoint yields inaccurate temporal differences and noisy GAE estimates until it tracks the return, and this recovery is slower on tool use trajectories.
Once the critic tracks the return, single-rollout GAE assigns credit along long tool interactions, while single-turn reasoning gains little beyond a sequence-level baseline.

These findings motivate \textbf{\liisa}, which \textbf{Li}fts \textbf{Li}ght \textbf{S}ingle-rollout \textbf{A}synchronous \textbf{O}ptimization through group relative actor-critic bootstrapping before transitioning to single rollout optimization.
During critic cold start, \liisa samples a small response group per prompt, trains the actor with normalized group relative advantages, and trains the critic on group centered residual returns.
Once the critic is deemed sufficiently calibrated, \liisa switches once to single rollout GAE.
Our policy gradient analysis explains how group comparisons align early updates with success probability and how a calibrated critic lets single-rollout GAE recover that direction.
Experiments on reasoning and agentic tasks show that \liisa catches up with or surpasses GRPO in early training and further raises the later performance ceiling.
Our contributions are summarized as follows:
\begin{itemize}
    \item \textbf{Lightweight Reproduction.} We implement in-flight asynchronous RL on \textsc{slime} and reproduce \lisao without the inaccessible scaled value pretraining stage for single-turn reasoning and multi-turn agentic tasks.
    \item \textbf{Cold-critic Diagnosis.} We conduct a controlled comparison of GRPO and \lisao across both task families. Our study reveals their distinct strengths, identifies the cold critic as the source of the early \lisao gap, and provides insights into group based and value based asynchronous optimization.
    \item \textbf{Group-relative Lifting.} We propose \liisa, which uses group-relative actor-critic bootstrapping and then switches once to single rollout GAE. Our theoretical analysis and empirical evidence showing that this design improves early optimization while retaining the later benefits of temporal credit assignment.
\end{itemize}

\section{Preliminaries}
\label{sec:prelim}

In reinforcement learning for language models, let \(q \sim \mathcal{D}\) denote a prompt and let \(\pi_\theta\) denote the policy.
A trajectory \(\tau\) is generated by \(\pi_\theta\) interacting with an environment.
For a single-turn task, \(\tau\) contains only one model response.
For a multi-turn agentic task, model-generated actions are interleaved with observation sequences returned by the environment.
We serialize all action and observation tokens after \(q\) as \(z_{1:T}\) and define the state \(s_t=(q,z_{<t})\).
We set \(m_t=1\) when \(z_t\) is generated by the model and \(m_t=0\) when it belongs to an environment observation.
At a model-generated position, we write \(a_t=z_t\), so \(a_t\) is the next model-generated token sampled from \(\pi_\theta(\cdot\mid s_t)\).
Policy loss, value loss, and advantage estimation are restricted to positions where \(m_t=1\).
The environment assigns a trajectory reward \(R(\tau)\) at the end of the interaction.
We write \(r_t\) for the reward assigned to model-generated token \(a_t\).
In this work, \(r_t=0\) except at the final model-generated position, where it equals \(R(\tau)\).

\subsection{In-Flight Asynchronous RL}
\label{sec:prelim_inflight}

In synchronous RL, rollout workers generate a complete batch under a frozen policy, and the trainer starts an update only after the batch is complete
\citep{sheng-2025-hybridflow}.
Asynchronous RL instead runs generation and training concurrently
\citep{noukhovitch-2025-asynchronous}.
Completed trajectories enter a buffer immediately, and the trainer repeatedly consumes available trajectories while rollout workers continue generation.
If the trainer holds policy version \(v_l\), a trajectory produced by version \(v_s < v_l\) has \emph{policy lag} \(v_l-v_s\).
Large lag makes the data increasingly off-policy and can destabilize optimization
\citep{zheng2025prosperity,huang2026stableasync}.

In this paper, we study the \emph{in-flight} weight-update asynchronous RL setting implemented on \textsc{slime}.
During a weight update, an unfinished rollout is paused without discarding its generated prefix and, after synchronization, resumes from that prefix under the latest policy.
Consequently, a model-generated token \(a_t\) may be sampled from a rollout policy
\(\pi_{\mathrm{r},v_t}\), where the version \(v_t\) can change within the same trajectory.
The RL system we study stores \(\log\pi_{\mathrm{r},v_t}(a_t \mid s_t)\) in the rollout engine during generation.
Thus, for the current policy \(\pi_\theta\), the token-level importance ratio can be computed as \(\rho_t(\theta)=\frac{\pi_\theta(a_t \mid s_t)}{\pi_{\mathrm{r},v_t}(a_t \mid s_t)}\).
Given an advantage estimate \(\hat A_t\), the clipped policy objective is
\begin{equation}
    \mathcal{J}(\theta)
    =
    \mathbb{E}_{q\sim\mathcal{D},\,\tau\sim(\pi_{\mathrm{r}},\mathcal{E})}
    \left[
        \frac{1}{\sum_{t=1}^{T}m_t}
        \sum_{t=1}^{T}m_t\ell_t(\theta)
    \right],
    \label{eq:clipped_objective}
\end{equation}
where \(\mathcal{E}\) is the environment, \(\pi_{\mathrm{r}}\) denotes the (possibly version-changing) rollout policy, and \(\ell_t(\theta)=\min\bigl(\rho_t(\theta)\hat A_t,\allowbreak \operatorname{clip}(\rho_t(\theta),1-\epsilon_\ell,1+\epsilon_h)\hat A_t\bigr)\).
After an update, leftover trajectories may remain in the rollout buffer and become increasingly off-policy.
In this paper, we cap this staleness with a maximum policy lag \(k\): samples whose lag exceeds this threshold are discarded.
Different RL algorithms mainly differ in how they estimate \(\hat A_t\) and how they handle extreme importance ratios.

\subsection{Group-wise Sampling and GRPO}
\label{sec:prelim_grpo}

GRPO avoids learning a value function by constructing a sampled baseline from the rewards of multiple trajectories for each prompt
\citep{shao-2024-deepseekmath}.
Given a prompt \(q\) and a group of \(n\) trajectories
\(\{\tau_i\}_{i=1}^{n}\) with rewards \(\{R_i\}_{i=1}^{n}\),
the group-relative advantage for trajectory \(i\) is
\begin{equation}
    \hat A_i^{\mathrm{GRPO}}
    =
    \frac{R_i-\frac{1}{n}\sum_{j=1}^{n}R_j}
    {\sqrt{\frac{1}{n-1}\sum_{j=1}^{n}\bigl(R_j-\frac{1}{n}\sum_{k=1}^{n}R_k\bigr)^2}+\varepsilon}.
    \label{eq:grpo_advantage}
\end{equation}
This scalar is assigned to every model-generated token in trajectory \(i\).
Equal group rewards yield zero advantage and no policy gradient.

GRPO offers a strong critic-free estimator when multiple comparable samples can be collected efficiently.
However, computing group statistics requires waiting for the slowest trajectory, creating a prompt-level synchronization barrier that increases policy lag.
In multi-turn trajectories, the sequence-level advantage in Eq.(\ref{eq:grpo_advantage}) also assigns the same advantage to every reasoning step and tool call.

\subsection{Single-rollout Optimization and SAO}
\label{sec:prelim_sao}

Single rollout optimization sets \(n=1\) and removes the need to assemble prompt level groups.
Since the group-relative advantage degenerates, it learns a critic \(V_\phi(s_t)\) that predicts the expected return from state \(s_t\).
PPO commonly uses GAE to construct token-level advantages
\citep{schulman-2016-gae,schulman-2017-ppo}.
For adjacent model-generated positions \(t\) and \(t+1\), the temporal-difference (TD) residual and GAE recursion are
\begin{gather}
    \delta_t
    =
    r_t+\gamma V_\phi(s_{t+1})-V_\phi(s_t),
    \label{eq:td_residual}
    \\
    \hat A_t^{\mathrm{GAE}}
    =
    \delta_t+\gamma\lambda\hat A_{t+1}^{\mathrm{GAE}},
    \label{eq:gae_recursion}
\end{gather}
where \(\gamma\) is the discount factor and \(\lambda\) controls the bias and variance tradeoff.
The corresponding return target is
\(\hat G_t=\hat A_t^{\mathrm{GAE}}+V_\phi(s_t)\), which the critic fits by minimizing
\begin{equation}
    \mathcal{L}_{V}(\phi)
    =
    \mathbb{E}_t\left[
        m_t\left(V_\phi(s_t)-\hat G_t\right)^2
    \right].
    \label{eq:value_loss}
\end{equation}
Unlike the trajectory-level GRPO advantage in Eq.(\ref{eq:grpo_advantage}), GAE produces token-level advantages that can vary across model-generated positions within a trajectory.
Its quality, however, directly depends on the accuracy of \(V_\phi\).

Agentic trajectories contain observation tokens that are produced by the environment \(\mathcal{E}\) rather than sampled from \(\pi_{\mathrm{r}}\).
SAO~\citep{hou-2026-sao} handles them with skip observation GAE.
Let \(u_1<\cdots<u_M\) be the positions for which \(m_t=1\), where \(M=\sum_{t=1}^{T}m_t\).
The skip observation GAE recursion is evaluated only on this subsequence as
\begin{gather}
    \delta_{u_j}
    =
    r_{u_j}
    +
    \gamma V_\phi(s_{u_{j+1}})
    -
    V_\phi(s_{u_j}),
    \label{eq:skip_obs_td}
    \\
    \hat A_{u_j}
    =
    \delta_{u_j}
    +
    \gamma\lambda\hat A_{u_{j+1}},
    \label{eq:skip_obs_gae}
\end{gather}
for \(j<M\).
At the terminal model-generated token, zero bootstrapping gives
\(\delta_{u_M}=r_{u_M}-V_\phi(s_{u_M})\) and
\(\hat A_{u_M}=\delta_{u_M}\).
Observation positions receive no policy or value loss.
This links consecutive model-generated tokens, including tokens separated by an observation sequence, while avoiding value transitions between externally generated tokens.

\begin{figure*}[t!]
  \centering
  \includegraphics[width=\textwidth]{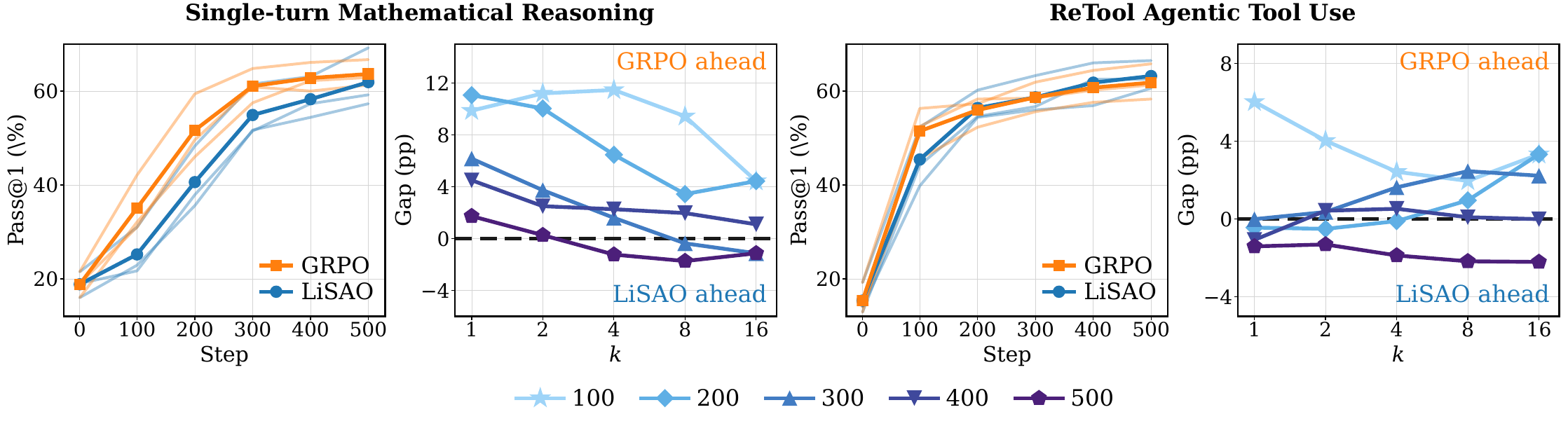}
  \vspace{-5mm}
  \caption{
  \textbf{GRPO vs. \lisao} on single-turn mathematical reasoning (left) and ReTool agentic tool use (right).
  Each half shows Pass@1 across training steps and Gap (pp) across Pass@$k$.
  Gap (pp) is GRPO minus \lisao, and a positive value means GRPO is ahead.
  Bold curves average AIME 2024, 2025, and 2026. Thin curves are the individual years.
  \textbf{The complete per-year pass@$k$ results are in Figure~\ref{fig:grpo_vs_lisao_full} of Appendix~\ref{app:grpo-vs-lisao-full}.}
  }
  \label{fig:grpo_vs_lisao}
  \end{figure*}

SAO further applies direct double-sided importance sampling (DIS), which forms \(\rho_t(\theta)\) from the stored rollout probabilities and masks tokens whose ratios fall outside a prescribed interval:
\begin{gather}
    \ell_t^{\mathrm{DIS}}(\theta)
    =
    f\bigl(\rho_t(\theta);\,\epsilon_\ell,\epsilon_h\bigr)
    \hat A_t,
    \label{eq:dis_loss}
    \\
    f(x;\epsilon_\ell,\epsilon_h)
    =
    \begin{cases}
        x, & 1-\epsilon_\ell < x < 1+\epsilon_h, \\
        0, & \text{otherwise}.
    \end{cases}
    \label{eq:dis}
\end{gather}

Unlike the PPO clip in Eq.(\ref{eq:clipped_objective}), out-of-range tokens receive a zero multiplier and do not enter the policy gradient.
However, SAO~\citep{hou-2026-sao} identifies cold-start value estimation as a major bottleneck and finds scaled value pretraining essential for robustly initializing \(V_\phi\) during early-stage optimization.
Without that initialization, the critic begins uncalibrated and the resulting GAE errors become the central challenge in this work.

\section{LiSAO without Value Pretraining}
\label{sec:lisao}

In this section, we empirically test whether SAO remains effective without scaled value pretraining.
We compare GRPO and \lisao under the same in-flight asynchronous pipeline on \textsc{slime}.
GRPO estimates a sequence-level advantage from multiple rollouts of the same prompt~\citep{shao-2024-deepseekmath}.
\lisao is our lightweight reproduction of SAO~\citep{hou-2026-sao} and uses the same single-rollout update, DIS, and skip-observation GAE.
Its critic is a separate network initialized from the policy checkpoint, with attention frozen and only the MLP blocks and a scalar value head trained.
We train Qwen3-4B~\citep{yang-2025-qwen3} for 500 steps on DAPO-Math-17K~\citep{yu-2025-dapo} with a binary outcome reward and a 32K context window, and disable thinking throughout.
The comparison covers single-turn mathematical reasoning, which starts from the base checkpoint and generates a text solution, and multi-turn agentic tool use under ReTool~\citep{feng-2025-retool}, which starts from a tool-use supervised checkpoint and interleaves reasoning with a Python interpreter.
We evaluate on AIME 2024, AIME 2025, and AIME 2026~\citep{maa-aime}, drawing 16 samples per problem at temperature $1.0$ and reporting the unbiased pass@$k$ estimator of \citet{chen-2021-codex}.
Experimental details are in Appendix~\ref{app:exp-details}, and the choices that distinguish this reproduction from the original SAO pipeline are in Appendix~\ref{app:lisao-repro}.

\subsection{Phenomenon: A Shared Early Gap Precedes Complementary Strengths}
 
\paragraph{GRPO leads on single-turn mathematical reasoning.}
The left half of Figure~\ref{fig:grpo_vs_lisao} follows this comparison across training steps and across $k$.
GRPO leads in pass@1 for most of training, and that lead then closes.
Averaged over AIME 2024, 2025, and 2026, the margin is 9.9\% at step 100 and 11.1\% at step 200, and it falls to 1.7\% at step 500.
At that checkpoint \lisao is ahead by 2.5\% on AIME 2024.
Drawing more samples narrows the same lead.
Averaged over the three years and all five checkpoints, the margin drops from 6.7\% at pass@1 to 1.5\% at pass@16.
The step 100 checkpoint on AIME 2024 is the exception, where the margin widens as $k$ grows, as shown in Figure~\ref{fig:grpo_vs_lisao_full}.

\paragraph{\lisao reaches a higher ceiling on multi-turn agentic tool use.}
The right half of Figure~\ref{fig:grpo_vs_lisao} gives the ReTool results in the same two panels.
GRPO leads in pass@1 early, and \lisao then moves ahead.
Averaged over AIME 2024, 2025, and 2026, \lisao trails by 6.0\% at step 100 and leads by 1.4\% at step 500.
At that checkpoint it is ahead on all three years.
This final lead widens as more samples are drawn.
At step 500, averaged over the three years, the margin rises from 1.4\% at pass@1 to 2.2\% at pass@16.
More details are in Figure~\ref{fig:grpo_vs_lisao_full}.

\begin{figure}[t]
    \centering
    \includegraphics[width=\linewidth]{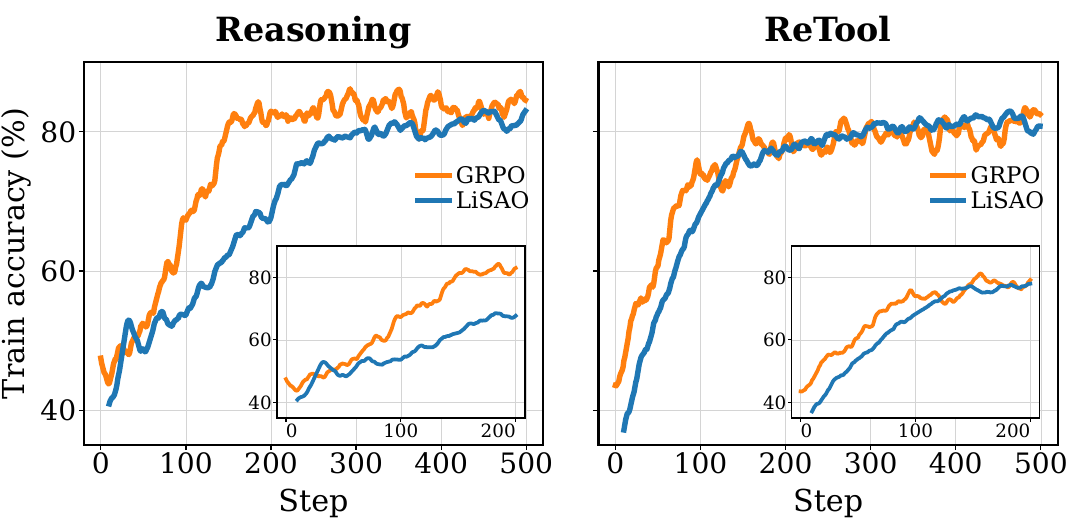}
    \caption{
    \textbf{Training accuracy} of GRPO and \lisao.
    Curves are a two-sided EMA (span 10) of per-step 0/1 accuracy.
    The inset shows the first 200 steps.
    \lisao starts at step 10, after critic warmup.
    }
    \label{fig:early_train_reward}
\end{figure}

\paragraph{A shared early-training deficit.}
Despite their complementary late-stage strengths, \lisao substantially trails GRPO over the early steps in both task families, as shown in panels 2 and 4 of Figure~\ref{fig:grpo_vs_lisao}.
Figure~\ref{fig:early_train_reward} shows that this deficit is already present on the training batches: \lisao's binary accuracy remains below GRPO through roughly step 200 in both single-turn math and ReTool. On single-turn math, the two methods start with similar training accuracy, but \lisao begins to fall behind after roughly 50 steps and the gap subsequently widens.
On ReTool, by contrast, \lisao starts with a substantial deficit during the first 100 steps, after which the gap gradually narrows.
Thus, although GRPO holds an early advantage in both settings, that advantage emerges at different stages of training.
As the next subsection shows, this delayed takeoff stems from the cold critic: single-rollout GAE becomes competitive only after the critic learns useful value estimates.

\subsection{Mechanism: Critic Dynamics Explain the Early Gap and GAE's Informative Turn}

\begin{figure}[t]
    \centering
    \includegraphics[width=\linewidth]{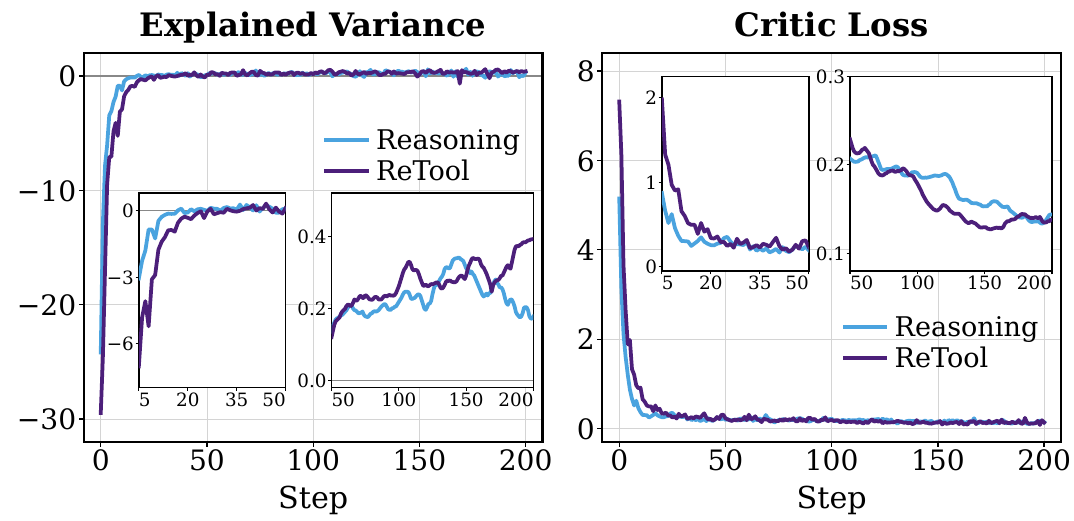}
    \caption{
    \textbf{Critic Dynamics} of \lisao over the first 200 steps.
    Insets magnify steps 5 to 50 and steps 50 to 200 to show the cold-start recovery and the later positive fit.
    The critic is updated from step 0, including 10 warmup steps before actor updates.
    }
    \label{fig:critic_dynamics}
\end{figure}

\paragraph{A cold critic makes early GAE noisy.}
Without value pretraining, \(V_\phi\) is initialized from the policy checkpoint before it predicts returns.
Explained variance, \(1-\operatorname{Var}_t(\hat G_t-V_\phi(s_t))/\operatorname{Var}_t(\hat G_t)\), is then negative, so the critic is worse than a predictor of the mean of \(\hat G_t\).
Figure~\ref{fig:critic_dynamics} shows the fit over the first 200 steps.
At step 10, when actor updates begin, explained variance is still about $-3.0$ on reasoning and about $-5.0$ on ReTool, with value loss about $0.9$ and $1.5$.
On reasoning, explained variance reaches about $0.0$ by step 30, and value loss falls below $0.4$ by step 20.
On ReTool, explained variance is still about $-1.0$ at step 20 and crosses $0.0$ around step 36.
Intermediate rewards are zero, so the residual in Eq.(\ref{eq:skip_obs_td}) reduces to \(\gamma V_\phi(s_{u_{j+1}})-V_\phi(s_{u_j})\), and Eq.(\ref{eq:skip_obs_gae}) accumulates that error into the advantage.
Early actor updates therefore follow critic noise, and this noise lasts longer on ReTool, where the training accuracy deficit appears immediately.

\paragraph{Return tracking separates the two tasks.}
After this recovery, value loss keeps falling on both tasks, and explained variance stays positive.
On reasoning, one terminal reward leaves little structure among tokens inside a completed solution, and a group of solutions already supplies a sequence-level baseline.
Training accuracy falls behind GRPO after about step 50, when explained variance is already positive.
On ReTool, the terminal reward folds an early plan, a tool call, and a later mistake into one scalar.
Skip-observation GAE separates those turns once \(V_\phi\) tracks the return.
Explained variance there rises from about $0.1$ at step 50 to about $0.4$ at step 200.
The accuracy gap narrows over the same window, and \lisao later moves ahead.
These dynamics motivate training the actor with a group-relative advantage and the critic with a group-centered return until \(V_\phi\) is ready, then switching to single-rollout GAE.

\section{\texorpdfstring{\liisa: Lifting \lisao via Group- Relative Actor-Critic Bootstrapping}{Li2SAO: Lifting LiSAO via Group-Relative Actor-Critic Bootstrapping}}
\label{sec:method}

\subsection{Bootstrapping Actor and Critic}
\label{sec:bootstrap}
\liisa addresses the cold critic without a separate value pretraining stage.
It begins with \(n_b>1\) responses for each prompt and jointly bootstraps the actor and critic from the resulting group.
Let \(g\) denote a prompt group and let \(R_i\) be the raw outcome reward of response \(i\).
We define the group mean \(\mu_g=\frac{1}{n_b}\sum_{j\in g}R_j\) and standard deviation \(\sigma_g=\sqrt{\frac{1}{n_b-1}\sum_{j\in g}(R_j-\mu_g)^2}\).

\paragraph{Group relative actor update.}
During this bootstrap phase, the actor uses \(\hat A_{i,t}^{\mathrm{actor}}=\frac{R_i-\mu_g}{\sigma_g+\varepsilon}\) at every model generated token in response \(i\).
This is the GRPO advantage in Eq.(\ref{eq:grpo_advantage}), and these advantages enter the clipped policy objective in Eq.(\ref{eq:clipped_objective}).
The importance-ratio correction for in-flight samples otherwise follows \lisao\ in Eq.(\ref{eq:dis_loss}) and Eq.(\ref{eq:dis}).

\paragraph{Group residual critic target.}
In vanilla LiSAO, Eq.(\ref{eq:value_loss}) fits \(V_\phi(s_{i,t})\) to the absolute return \(\hat G_{i,t}\) built from the outcome reward \(R_i\) on the final token.
Inside a prompt group, the outcome in that return splits as \(R_i=\mu_g+(R_i-\mu_g)\), where \(\mu_g\) is identical for every response and reflects only how easy the current prompt is.
Predicting this shared mean from the prompt lowers Eq.(\ref{eq:value_loss}) even when a successful response and an unsuccessful response receive the same value.
Thus, \liisa sets \(\widetilde G_{i,t}=\hat G_{i,t}-\mu_g\) and drops groups with identical rewards, where \(R_i-\mu_g=0\), so the squared loss supervises the critic by which response in the group succeeds.

\begin{figure}[!t]
    \centering
    \includegraphics[width=\linewidth]{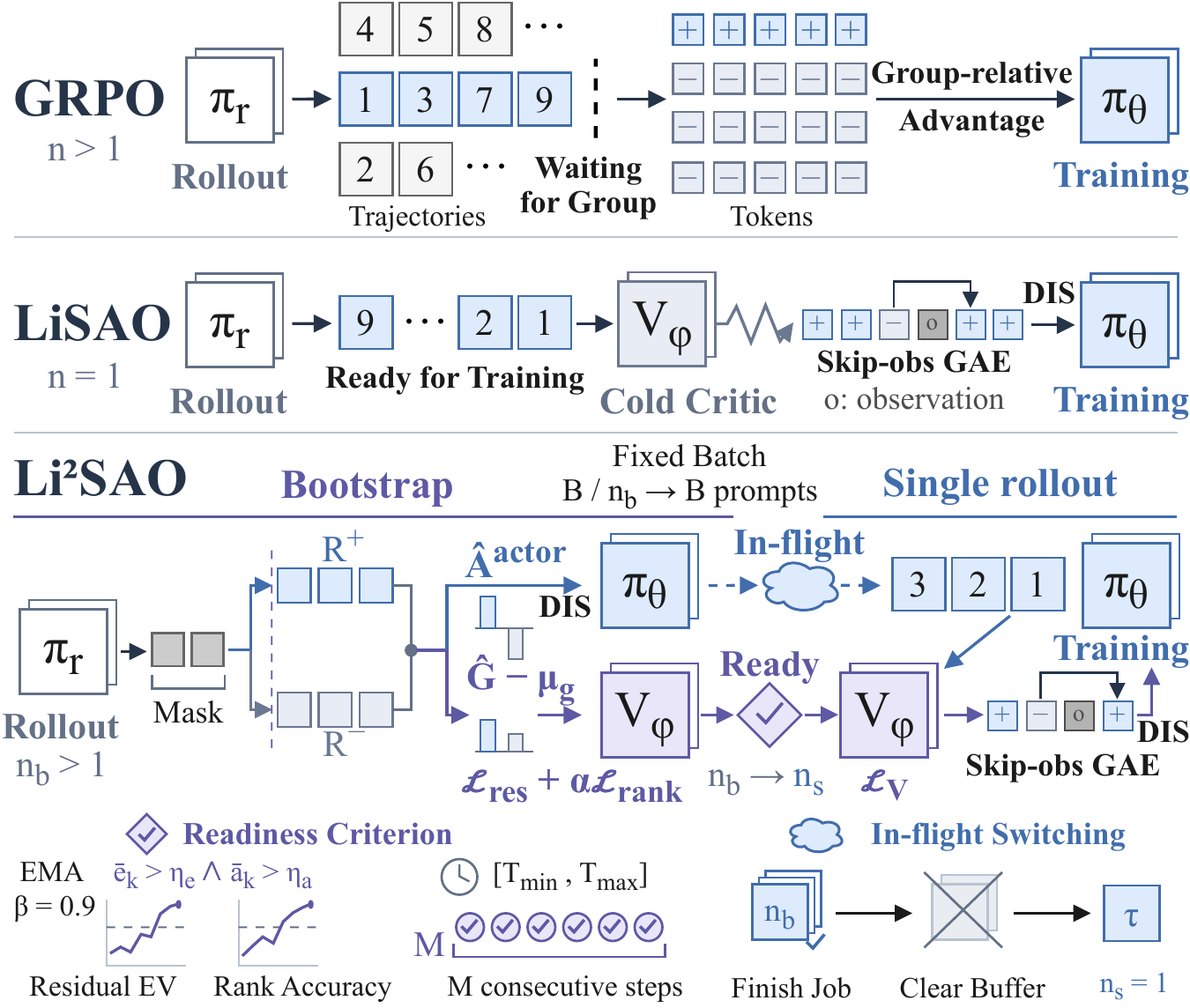}
    \caption{
    \textbf{Overview of \liisa.}
    GRPO waits for a complete prompt group and assigns one group-relative advantage to every action token.
    \lisao consumes each rollout on arrival, but a cold critic makes early GAE noisy.
    \liisa bootstraps the actor with group-relative advantages and the critic with centered returns, divergence masking, and terminal ranking, then switches to single-rollout GAE when residual explained variance and ranking accuracy exceed the thresholds.
    }
    \label{fig:li2sao_method}
\end{figure}

\paragraph{Divergence masked residual loss.}
Responses with different outcomes can share a common prefix.
\liisa does not train \(V_\phi\) on states before those responses diverge, because a residual target there depends on future tokens that the state has not generated.
For response \(i\), let \(d(i,j)\) be the first response position at which it differs from response \(j\), and set \(d_i=\max_{j\in g:R_j\ne R_i}d(i,j)\).
\liisa sets the critic mask \(c_{i,t}=1\) on model generated tokens at and after \(d_i\).
When the local training partition contains no response with \(R_j\ne R_i\), \liisa instead keeps only the last action token.
With this mask, \liisa freezes \(V^{\mathrm{old}}_{i,t}=V_{\phi_{\mathrm{old}}}(s_{i,t})\) once before the critic passes on a batch.
Writing \(V_{i,t}=V_\phi(s_{i,t})\) and \(\epsilon_v=0.2\), it minimizes
\begin{equation}
    \begin{gathered}
        \bar V_{i,t}
        =
        V^{\mathrm{old}}_{i,t}
        +
        \operatorname{clip}(V_{i,t}-V^{\mathrm{old}}_{i,t},-\epsilon_v,\epsilon_v),
        \\
        \mathcal{L}_{\mathrm{res}}(\phi)
        =
        \frac{
            \sum_{g,i,t}
            c_{i,t}
            \max\left\{
            \begin{gathered}
                (V_{i,t}-\widetilde G_{i,t})^2,\\
                (\bar V_{i,t}-\widetilde G_{i,t})^2
            \end{gathered}
            \right\}
        }{
            \sum_{g,i,t} c_{i,t}
        }.
    \end{gathered}
    \label{eq:li2sao_residual_loss}
\end{equation}

\paragraph{Terminal ranking loss.}
Eq.(\ref{eq:li2sao_residual_loss}) fits each kept token to its own centered return, but it contains no comparison between responses.
Reducing that loss on one completion can leave \(V_\phi(s_{i^+,T_{i^+}})\le V_\phi(s_{i^-,T_{i^-}})\).
The return on the last action equals the outcome reward, so for a mixed pair with \(R_{i^+}>R_{i^-}\) the centered targets of the two states differ by \(R_{i^+}-R_{i^-}\).
Let \(\mathcal{P}\) be the set of such pairs, we therefore add the terminal ranking loss:
\begin{equation}
    \begin{aligned}
        \mathcal{L}_{\mathrm{rank}}(\phi)
        &=
        \frac{1}{|\mathcal{P}|}
        \sum_{(i^+,i^-)\in\mathcal{P}}
        \bigl[
            m
        \\
        &\quad
            -V_\phi(s_{i^+,T_{i^+}})
            +V_\phi(s_{i^-,T_{i^-}})
        \bigr]_+.
    \end{aligned}
    \label{eq:li2sao_rank_loss}
\end{equation}
Then critic loss is
\(\mathcal{L}_{\mathrm{critic}}
=\mathcal{L}_{\mathrm{res}}+\alpha\mathcal{L}_{\mathrm{rank}}\).
We use margin \(m=0.1\) and coefficient \(\alpha=0.1\).
Actor and critic optimization begin together at the first step, so \liisa requires no critic-only warmup.

\subsection{From Group Sampling to Single Rollout}
Group sampling supplies a reliable actor signal while the critic learns to distinguish outcomes.
Once that distinction is reliable, retaining \(n_b\) responses per prompt is unnecessary and delays asynchronous updates.
\liisa therefore makes a single irreversible transition from \(n_b\) to \(n_s=1\).
The total number of trajectories per update stays fixed, so the number of prompts increases from \(B/n_b\) to \(B\).
After the transition, the actor uses GAE in Eq.(\ref{eq:gae_recursion}) and the critic resumes fitting absolute returns in Eq.(\ref{eq:value_loss}).
The residual transformation and ranking loss become inactive because singleton groups contain no within group comparison.

\paragraph{Readiness criterion.}
We measure critic readiness on mixed groups using the centered terminal outcome
\(Z_i=R_i-\mu_g\).
At bootstrap step \(k\), the residual explained variance is
\begin{equation}
    e_k
    =
    1-
    \frac{
        \operatorname{Var}_i
        \left[
            Z_i-V_\phi(s_{i,T_i})
        \right]
    }{
        \operatorname{Var}_i[Z_i]
    }.
    \label{eq:li2sao_residual_ev}
\end{equation}
We also compute the fraction of cross outcome pairs ordered correctly:
\begin{equation}
        a_k=
        \frac{1}{|\mathcal{P}|}
        \sum_{(i^+,i^-)\in\mathcal{P}}
        \mathbb{I}
        \bigl[
            V_\phi(s_{i^+,T_{i^+}})
            >
            V_\phi(s_{i^-,T_{i^-}})
        \bigr].
    \label{eq:li2sao_rank_accuracy}
\end{equation}
Both statistics are aggregated across data parallel workers and smoothed with exponential moving averages
\(\bar e_k=\beta\bar e_{k-1}+(1-\beta)e_k\) and
\(\bar a_k=\beta\bar a_{k-1}+(1-\beta)a_k\).
The critic is ready when \(\bar e_k>\eta_e\) and \(\bar a_k>\eta_a\) for \(M\) consecutive steps after a minimum bootstrap duration \(T_{\min}\).
A maximum duration \(T_{\max}\) provides a fallback when the gate does not open.
Our main configuration uses
\(\beta=0.9\),
\(\eta_e=0.2\),
\(\eta_a=0.65\),
\(M=10\),
\(T_{\min}=60\), and
\(T_{\max}=150\).
The one way transition prevents noisy batch level fluctuations from returning training to group sampling.

\paragraph{In-flight switching.}
In in-flight asynchronous RL system, generation overlaps training, so one job is already prefetched when the gate opens.
\liisa finishes that job at group size \(n_b\), discards leftover bootstrap groups still held by the rollout worker, and only then sets the group size to \(n_s=1\).
The next job contains singleton groups, so every batch has one group size and one advantage estimator.

\subsection{Theoretical Analysis}
\label{sec:theory}
To provide insight into each \liisa stage, we characterize its expected update under the policy gradient.
Fix a prompt, a differentiable policy \(\pi=\pi_\theta\), a binary terminal reward \(R\), and discount \(\gamma=1\).
The state \(s_t\) before model action \(a_t\) contains only the preceding history, including tool observations.
Let \(p=\mathbb E_\pi R\) be the success probability of this prompt, and let \(V^\pi(s)\) be the expected remaining return from state \(s\).
Write \(\psi_t=\nabla_\theta\log\pi_\theta(a_t\mid s_t)\) and \(S=\sum_t\psi_t\), where the sum runs over model actions.
The first result characterizes the group-bootstrap signal and its residual critic target.
Regularity conditions and proofs appear in Appendix~\ref{app:theory}.

\begin{proposition}
\label{prop:group-identification}
Sample \(n=n_b\ge2\) independent trajectories from \(\pi\), with \(0<p<1\).
Set \(Y_i=R_i-\mu_g\) and \(\mathcal M=\{0<\sum_iR_i<n\}\).
Define the reference group score as
\(\widehat g_{\mathrm{grp}}=n^{-1}\sum_i S_iY_i/(\sigma_g+\varepsilon)\), where \(\varepsilon>0\), then its expectation is a positive rescaling of the success-probability gradient:
\begin{equation}
 \mathbb E\widehat g_{\mathrm{grp}}
 =\omega_n(p)\nabla_\theta p,
 \quad \omega_n(p)>0.
\label{eq:theory-group-gradient}
\end{equation}
At a pre-action state \(s\), the optimal target for the base squared residual loss on mixed groups is
\begin{equation}
\begin{gathered}
 F^\star(s)
=\mathbb E[Y_i\mid s_{i,t}=s,\mathcal M]\\
=\frac{(1-1/n)(V^\pi(s)-p)}
        {1-p^{n-1}V^\pi(s)-(1-p)^{n-1}(1-V^\pi(s))}.
\end{gathered}
\label{eq:theory-residual-target}
\end{equation}
For fixed \(p\), \(F^\star(s)\) strictly increases with \(V^\pi(s)\).
\end{proposition}
Thus, this characterization isolates what group centering asks the critic to learn: group-residual supervision prevents the critic from reducing its regression loss merely by predicting prompt difficulty, and instead preserves response-dependent distinctions relevant to relative success.
The next proposition shows exact conditional values turn TD residuals and GAE into single-trajectory estimators of the success-probability gradient.

\begin{proposition}
\label{prop:single-reconstruction}
Let \(\delta_t^\pi=r_t+V^\pi(s_{t+1})-V^\pi(s_t)\) be the temporal-difference residual of the exact conditional value \(V^\pi\).
For any \(\lambda\in(0,1]\), this residual and the fixed-trace GAE are single-trajectory estimators of the success-probability gradient:
\begin{equation}
 \nabla_\theta p
 =\mathbb E_\pi\!\left[\sum_t\psi_t\delta_t^\pi\right]
 =\mathbb E_\pi\!\left[\sum_t\psi_t\widehat A_t^\lambda(V^\pi)\right].
\label{eq:theory-single-gradient}
\end{equation}
\end{proposition}

This proposition specializes the policy-gradient and GAE identities~\citep{sutton-1999-policy-gradient,schulman-2016-gae} to the model-action history.
Thus, the identity identifies the ideal target of the switch: at \(V^\pi\), single-rollout GAE keeps the group bootstrap's expected direction.
\textbf{More theoretical analyses about off-policy estimation, value calibration, and task structure are in Appendix~\ref{app:theory}.}

\section{Experiments}
\label{sec:exp}

\begin{table*}[t]
\centering
\small
\setlength{\tabcolsep}{2.6pt}
\renewcommand{\arraystretch}{1.12}
\caption{
\textbf{Pass@$k$} (\%) at step 500 on AIME 2024, 2025, and 2026.
Base and SFT are the shared checkpoints before RL: single-turn reasoning starts from the base model and ReTool from a tool-use supervised checkpoint.
All rows use 16 samples per problem at temperature 1.0 and the unbiased estimator of \citet{chen-2021-codex}.
Avg. is the mean over the three years.
The best results are highlighted in \textbf{Bold}.
}
\label{tab:main-results}
\begin{tabular}{@{} l *{5}{c} @{\hspace{0.7em}} *{5}{c} @{\hspace{0.7em}} *{5}{c} @{\hspace{0.7em}} *{5}{c} @{}}
\toprule
& \multicolumn{5}{c}{AIME 2024}
& \multicolumn{5}{c}{AIME 2025}
& \multicolumn{5}{c}{AIME 2026}
& \multicolumn{5}{c}{Avg.} \\
\cmidrule(lr){2-6} \cmidrule(lr){7-11} \cmidrule(lr){12-16} \cmidrule(lr){17-21}
Pass@$k$
& 1 & 2 & 4 & 8 & 16
& 1 & 2 & 4 & 8 & 16
& 1 & 2 & 4 & 8 & 16
& 1 & 2 & 4 & 8 & 16 \\
\midrule
\multicolumn{21}{@{}l@{}}{\cellcolor{bandblue}\rule[-0.42ex]{0pt}{2.35ex}\textit{Single-turn Mathematical Reasoning}} \\
Base
& 21.5 & 29.3 & 36.7 & 45.6 & 56.7
& 19.0 & 24.7 & 29.5 & 34.3 & 40.0
& 16.0 & 22.4 & 30.2 & 40.0 & 50.0
& 18.8 & 25.5 & 32.1 & 40.0 & 48.9 \\
GRPO
& 66.7 & 74.4 & 79.2 & 81.5 & 83.3
& 61.3 & 69.9 & 74.4 & 78.4 & \textbf{83.3}
& 62.9 & 69.8 & 74.6 & 78.9 & 83.3
& 63.6 & 71.4 & 76.1 & 79.6 & 83.3 \\
\lisao
& 69.2 & 76.7 & 80.6 & 81.7 & 83.3
& 59.2 & 69.6 & \textbf{76.9} & \textbf{81.4} & \textbf{83.3}
& 57.3 & 67.0 & 74.4 & 80.9 & \textbf{86.7}
& 61.9 & 71.1 & 77.3 & 81.3 & 84.4 \\
\liisa
& \textbf{72.5} & \textbf{80.1} & \textbf{84.2} & \textbf{85.9} & \textbf{90.0}
& \textbf{65.0} & \textbf{72.3} & 76.7 & 79.8 & \textbf{83.3}
& \textbf{63.5} & \textbf{71.3} & \textbf{76.9} & \textbf{81.6} & \textbf{86.7}
& \textbf{67.0} & \textbf{74.6} & \textbf{79.3} & \textbf{82.4} & \textbf{86.7} \\
\midrule
\multicolumn{21}{@{}l@{}}{\cellcolor{bandblue}\rule[-0.42ex]{0pt}{2.35ex}\textit{ReTool Agentic Tool Use}} \\
SFT
& 19.2 & 31.9 & 47.3 & 61.0 & 66.7
& 14.0 & 22.3 & 32.4 & 41.5 & 50.0
& 12.9 & 22.0 & 34.1 & 48.0 & 60.0
& 15.4 & 25.4 & 37.9 & 50.2 & 58.9 \\
GRPO
& 65.8 & 75.6 & 82.1 & 87.5 & 90.0
& 58.3 & 67.1 & 73.3 & 77.2 & 80.0
& 61.3 & 71.2 & 79.4 & 84.6 & 86.7
& 61.8 & 71.3 & 78.3 & 83.1 & 85.6 \\
\lisao
& 66.5 & 75.5 & 83.1 & 88.2 & 90.0
& 60.6 & 69.3 & 75.7 & 80.4 & \textbf{83.3}
& 62.5 & 73.0 & 81.6 & 87.2 & 90.0
& 63.2 & 72.6 & 80.1 & 85.3 & 87.8 \\
\liisa
& \textbf{69.8} & \textbf{79.2} & \textbf{85.1} & \textbf{89.9} & \textbf{93.3}
& \textbf{61.0} & \textbf{69.7} & \textbf{77.5} & \textbf{80.5} & \textbf{83.3}
& \textbf{62.9} & \textbf{73.7} & \textbf{82.4} & \textbf{87.6} & \textbf{93.3}
& \textbf{64.6} & \textbf{74.2} & \textbf{81.7} & \textbf{86.0} & \textbf{90.0} \\
\bottomrule
\end{tabular}
\end{table*}

We evaluate \liisa on single-turn mathematical reasoning and multi-turn agentic tool use.
Our experimental design examines two questions: 
(i) whether group-relative actor-critic bootstrapping mitigates the early optimization deficit caused by a cold critic, and (ii) whether the subsequent switch to single-rollout GAE preserves temporal credit assignment and raises the later performance ceiling.

\paragraph{Experimental Setups.}
We compare \liisa with GRPO and \lisao under the training and evaluation protocol of Section~\ref{sec:lisao}.
\liisa follows the procedure in Section~\ref{sec:method}.
During the bootstrap, \liisa uses 8 responses per prompt, the same group size as GRPO, and begins the actor and critic updates together.
Further details are in Appendix~\ref{app:exp-details}.

\begin{figure}[t]
    \centering
    \includegraphics[width=\linewidth]{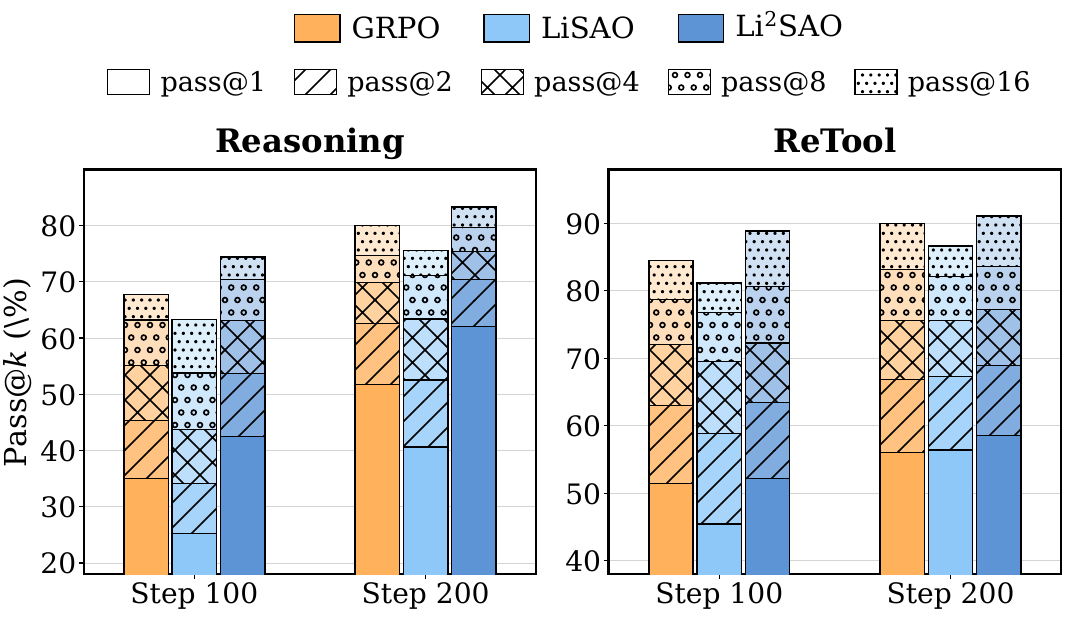}
    \vspace{-10pt}
    \caption{
    \textbf{Early pass@$k$} at steps 100 and 200.
    Each bar averages AIME 2024, 2025, and 2026.
    }
    \label{fig:early-pass1}
    \vspace{-3pt}
\end{figure}

\subsection{Main Results}

\paragraph{Group-relative bootstrapping reverses the early deficit.}
Figure~\ref{fig:early-pass1} shows that \liisa leads both baselines in average pass@$k$ at steps 100 and 200 across both task families.
At step 100, average pass@1 reaches 42.5\% on reasoning and 52.2\% on ReTool, exceeding \lisao by 17.3 and 6.8 percentage points and GRPO by 7.4 and 0.8 points, respectively.
The advantage at step 200 shows that these early gains persist beyond bootstrapping, supporting its effectiveness without value pretraining.

\paragraph{Single-rollout GAE raises the later performance ceiling.}
At step 500, \liisa achieves the highest average pass@$k$ for every evaluated $k$ in both settings (Table~\ref{tab:main-results}).
Average pass@1 reaches 67.0\% on reasoning and 64.6\% on ReTool, exceeding GRPO by 3.4 and 2.8 percentage points and \lisao by 5.1 and 1.4 points, respectively.
Average pass@16 also rises to 86.7\% and 90.0\%, indicating improved coverage under repeated sampling.
Thus, \liisa improves both early and final performance while retaining single-rollout GAE and its temporal credit assignment after bootstrapping.
Figure~\ref{fig:li2sao_full} reports the same comparison at each checkpoint and year.

\subsection{Additional Experimental Analysis}
To further analyze \liisa, we report additional experiments in Appendix~\ref{app:additional-analysis}, including \textbf{full per-year results} (Section~\ref{app:full-comparison}), \textbf{training dynamics} (Section~\ref{app:training-dynamics}), \textbf{critic-objective ablations} (Section~\ref{app:ablations}), \textbf{wall-clock efficiency} (Section~\ref{app:training-efficiency}), and \textbf{policy staleness} (Section~\ref{app:staleness-dynamics}).
We also evaluate \textbf{coding agentic RL} in Appendix~\ref{app:coding-agentic-rl}, training larger models on long-horizon software engineering.
These results demonstrate the effectiveness of \liisa.

\section{Conclusion}
\label{sec:concl}

This paper studies single-rollout asynchronous RL without scaled value pretraining and identifies a shared early optimization deficit associated with a cold critic, despite the complementary strengths of GRPO on reasoning and \lisao on agentic tasks.
Our method, \liisa, jointly bootstraps the actor with group-relative advantages and the critic with group-centered returns before switching once to single-rollout GAE.
The policy-gradient analysis explains the direction of group-based updates and its recovery by GAE under exact value estimation.
Experiments show that \liisa catches up with or surpasses GRPO in early training and further raises the later performance ceiling.
These findings support group-relative bootstrapping as a practical route to effective single-rollout asynchronous optimization without a pretrained value model.

\section*{Limitations}
\textbf{Theoretical Scope.} Our analysis is intended to isolate the mechanisms behind the two training stages.
Proposition~\ref{prop:group-identification} shows that group centered supervision removes the prompt difficulty component while preserving the ordering induced by conditional success probabilities.
Proposition~\ref{prop:single-reconstruction} then shows that, with exact conditional values, single rollout GAE recovers the same expected success direction.
These population identities support the method design but do not constitute a convergence guarantee for the complete training procedure.
The implemented critic also uses divergence masking, value clipping, and ranking regularization, so its optimum need not exactly equal \(F^\star\).

\noindent
\textbf{Experimental Scope.} Our experiments use a deliberately controlled setting based on one 4B parameter model family, DAPO Math 17K, binary outcome rewards, and AIME evaluation for both single turn reasoning and ReTool.
This design enables matched comparisons under a common asynchronous pipeline and training protocol, supporting a fine-grained empirical analysis of algorithmic behavior.
Because the scaled value model used by SAO is not publicly available, our comparison centers on \lisao, a transparent reproduction without value pretraining.



\bibliography{li2sao}

\appendix
\newpage
\onecolumn
\begingroup
\allowdisplaybreaks[1]
\section{From Group Comparisons to Single-Rollout Gradients}
\label{app:theory}

\subsection{Setup and conditioning conventions}
\label{app:theory-setup}
Fix a prompt \(q\) and a policy parameter \(\theta\), and write \(\pi=\pi_\theta\).
We use the undiscounted objective with \(\gamma=1\).
The index \(t\) counts model actions.
At each real decision, \(s_t\) contains the prompt, the decision index, and the complete history of actions, rewards, and tool observations available before \(a_t\).
For a fixed horizon \(H\), write \(\tau=(s_0,a_0,r_0,\ldots,s_{H-1},a_{H-1},r_{H-1},s_H)\).
The natural filtration before each action is
\begin{equation}
 \mathcal F_t
 =\sigma\!\left(s_0,(a_j,r_j,s_{j+1})_{0\le j<t}\right).
\end{equation}
In particular, \(a_t\), \(r_t\), and \(s_{t+1}\) are \(\mathcal F_{t+1}\)-measurable.
Let \(L\le H\) be the number of real actions, with \(L\ge1\) and \(\{L\le t\}\in\mathcal F_t\), so that \(L\) is a bounded stopping time.
After termination, pad the trajectory with an absorbing state, zero rewards, and a dummy action of probability one.
The filtration retains the observed history after termination.

The conditional distribution of \((s_{t+1},r_t)\) given \(\mathcal F_t\) and \(a_t\) does not depend directly on \(\theta\).
Model actions are discrete, and the probabilities \(\pi_\theta(a\mid s)\) are differentiable with locally fixed support.
We assume differentiation can be interchanged with the expectations below and that all displayed expectations are absolutely integrable.
Covariance statements additionally require finite second moments.
All trajectory laws and expectations are conditional on the fixed prompt.

Let \(R\in\{0,1\}\) be the reward observed at termination, and let \(p=\mathbb E_\pi R\).
Thus \(r_t=\mathbf1_{\{t=L-1\}}R\).
Define the return-to-go \(G_t=\sum_{k=t}^{H-1}r_k\) and the accumulated past reward \(R_{<t}=\sum_{k<t}r_k\), so that \(R=R_{<t}+G_t\).
The state value is \(V_t^\pi=V^\pi(s_t)=\mathbb E_\pi[G_t\mid\mathcal F_t]\), with zero value after termination.
Its temporal-difference residual is \(\delta_t^\pi=r_t+V_{t+1}^\pi-V_t^\pi\).
For any value function \(V\) that is zero at terminal and absorbing states, the generalized advantage estimator with fixed \(\lambda\in(0,1]\) is
\begin{equation}
 \widehat A_t^\lambda(V)
 =\sum_{k=t}^{H-1}\lambda^{k-t}
       \bigl(r_k+V(s_{k+1})-V(s_k)\bigr).
\label{eq:app-history-notation}
\end{equation}

Write \(\psi_t=\nabla_\theta\log\pi_\theta(a_t\mid s_t)\) for the policy score and \(S=\sum_{t=0}^{H-1}\psi_t\) for the trajectory score.
Dummy actions have zero score.
The conditional score has zero mean:
\begin{equation}
 \mathbb E_\pi[\psi_t\mid\mathcal F_t]
 =\sum_a\pi_\theta(a\mid s_t)\nabla_\theta\log\pi_\theta(a\mid s_t)
 =\nabla_\theta\sum_a\pi_\theta(a\mid s_t)
 =0.
\end{equation}
For any \(\mathcal F_t\)-measurable baseline \(B_t\) satisfying \(\mathbb E_\pi[\|\psi_t\|\,|B_t|]<\infty\), the tower property gives
\begin{equation}
 \mathbb E_\pi[\psi_tB_t]
 =\mathbb E_\pi\!\left[\mathbb E_\pi[\psi_tB_t\mid\mathcal F_t]\right]
 =\mathbb E_\pi\!\left[B_t\mathbb E_\pi[\psi_t\mid\mathcal F_t]\right]
 =0.
\end{equation}
Let \(\ell_\theta(\tau)\) denote the trajectory density with respect to a common reference measure \(\nu\) independent of \(\theta\).
Since only the action probabilities depend on \(\theta\), its score is
\begin{equation}
 \nabla_\theta\log\ell_\theta(\tau)
 =\sum_{t=0}^{H-1}\nabla_\theta\log\pi_\theta(a_t\mid s_t)
 =S.
\end{equation}
Differentiating the expected reward and using \(\mathbb E_\pi S=0\) yields the likelihood-ratio identity
\begin{equation}
 \nabla_\theta p
 =\int R(\tau)\nabla_\theta\ell_\theta(\tau)\,\nu(d\tau)
 =\mathbb E_\pi[RS]
 =\mathbb E_\pi[(R-p)S].
\label{eq:app-score-identity}
\end{equation}
The group calculation uses independent trajectories from this policy.
Critic regression conditions on a specified real state before any additional selection based on outcomes.

\subsection{Proof of Proposition~\ref{prop:group-identification}}
\label{app:group-proof}
\paragraph{Finite-group actor signal.}
The calculation holds for any integrable vector \(U=U(\tau)\) computed from one trajectory, with \(U_i=U(\tau_i)\).
Let \(K=\sum_iR_i\), \(K_{-i}=\sum_{j\ne i}R_j\), and \(Y_i=R_i-K/n\).
For a group containing \(k\) successes, write \(d_k=\sigma_k+\varepsilon\), where the sample variance is
\begin{equation}
 \sigma_k^2
 =\frac{k(1-k/n)^2+(n-k)(k/n)^2}{n-1}
 =\frac{k(n-k)}{n(n-1)}.
\end{equation}
With \(u_r=\mathbb E[U\mid R=r]\), define
\begin{equation}
 \widehat g(U)=\frac1n\sum_i\frac{U_iY_i}{d_K}.
\end{equation}
The pair \((U_i,R_i)\) is independent of \(K_{-i}\).
Thus, whenever the conditioning event has positive probability,
\begin{equation}
 \mathbb E[U_i\mid R_i=r,K=k]
 =\mathbb E[U_i\mid R_i=r,K_{-i}=k-r]
 =\mathbb E[U_i\mid R_i=r]=u_r.
\end{equation}
For \(1\le k\le n-1\), the binomial probabilities give
\begin{equation}
 \Pr(R_i=1\mid K=k)
 =\frac{\binom{n-1}{k-1}p^k(1-p)^{n-k}}
        {\binom nk p^k(1-p)^{n-k}}
 =\frac{k}{n}.
\end{equation}
Conditioning first on \(R_i\) therefore yields
\begin{equation}
\begin{aligned}
 \mathbb E\!\left[\sum_{i:R_i=1}U_i\,\middle|\,K=k\right]
 &=\sum_{i=1}^n\mathbb E[\mathbf1_{\{R_i=1\}}U_i\mid K=k]\\
 &=\sum_{i=1}^n\Pr(R_i=1\mid K=k)\mathbb E[U_i\mid R_i=1,K=k]\\
 &=\sum_{i=1}^n\frac{k}{n}u_1=ku_1.
\end{aligned}
\end{equation}
The unsuccessful trajectories similarly satisfy
\begin{equation}
 \mathbb E\!\left[\sum_{i:R_i=0}U_i\,\middle|\,K=k\right]
 =\sum_{i=1}^n\left(1-\frac{k}{n}\right)u_0
 =(n-k)u_0.
\end{equation}
Within each group, successful trajectories have residual \(1-k/n\), while unsuccessful trajectories have residual \(-k/n\).
Consequently,
\begin{equation}
\begin{aligned}
 \mathbb E[\widehat g(U)\mid K=k]
 &=\frac{1}{nd_k}\left[
       \left(1-\frac{k}{n}\right)
       \mathbb E\!\left[\sum_{i:R_i=1}U_i\,\middle|\,K=k\right]
       -\frac{k}{n}
       \mathbb E\!\left[\sum_{i:R_i=0}U_i\,\middle|\,K=k\right]
    \right]\\
 &=\frac{(1-k/n)ku_1-(k/n)(n-k)u_0}{nd_k}\\
 &=\frac{k(n-k)}{n^2d_k}(u_1-u_0).
\end{aligned}
\end{equation}
For \(k=0\) or \(k=n\), every \(Y_i\) is zero, so the group score vanishes.
For the remaining terms, use
\begin{equation}
 k(n-k)\binom nk
 =\frac{n!}{(k-1)!(n-k-1)!}
 =n(n-1)\binom{n-2}{k-1}.
\end{equation}
Since \(K\sim\operatorname{Binomial}(n,p)\), total expectation and the substitution \(z=k-1\) give
\begin{equation}
\begin{aligned}
 \mathbb E\widehat g(U)
 &=\sum_{k=1}^{n-1}\binom nk p^k(1-p)^{n-k}
       \frac{k(n-k)}{n^2d_k}(u_1-u_0)\\
 &=\frac{n-1}{n}(u_1-u_0)
       \sum_{k=1}^{n-1}\binom{n-2}{k-1}
         \frac{p^k(1-p)^{n-k}}{d_k}\\
 &=\frac{n-1}{n}p(1-p)(u_1-u_0)
       \sum_{k=1}^{n-1}\binom{n-2}{k-1}
         \frac{p^{k-1}(1-p)^{n-k-1}}{d_k}\\
 &=p(1-p)(u_1-u_0)\,
     \frac{n-1}{n}
     \sum_{z=0}^{n-2}\binom{n-2}{z}
       \frac{p^z(1-p)^{n-2-z}}{d_{z+1}}.
\end{aligned}
\end{equation}
Conditioning on the binary outcome also gives
\begin{equation}
 \mathbb E[U(R-p)]
 =p(1-p)u_1+(1-p)(-p)u_0
 =p(1-p)(u_1-u_0).
\end{equation}
Define
\begin{equation}
 \omega_n(p)
 :=\frac{n-1}{n}\sum_{z=0}^{n-2}\binom{n-2}{z}
       \frac{p^z(1-p)^{n-2-z}}{\sigma_{z+1}+\varepsilon}>0.
\end{equation}
Every summand is positive because \(0<p<1\), \(n\ge2\), and \(\varepsilon>0\).
The two expectations therefore satisfy
\begin{equation}
 \mathbb E\widehat g(U)=\omega_n(p)\mathbb E[U(R-p)].
\label{eq:app-group-general}
\end{equation}
Taking \(U=S\) gives \(\widehat g(U)=\widehat g_{\mathrm{grp}}\), and Eq.~\eqref{eq:app-score-identity} yields
\begin{equation}
 \mathbb E\widehat g_{\mathrm{grp}}
 =\omega_n(p)\mathbb E_\pi[S(R-p)]
 =\omega_n(p)\nabla_\theta p.
\end{equation}
This proves Eq.~\eqref{eq:theory-group-gradient}.

\paragraph{Residual conditional means and regression risk.}
The critic uses \(\lambda=1\) for its return target.
For a real decision \(t<L\), telescoping and zero terminal bootstrapping give
\begin{equation}
\begin{aligned}
 \widehat G_t
 &=V_\phi(s_t)+\sum_{k=t}^{L-1}
       \bigl(r_k+V_\phi(s_{k+1})-V_\phi(s_k)\bigr)\\
 &=\sum_{k=t}^{L-1}r_k+V_\phi(s_t)+V_\phi(s_L)-V_\phi(s_t)\\
 &=\sum_{k=t}^{L-1}r_k=R.
\end{aligned}
\end{equation}
Fix a real state \(s=s_{i,t}\) and put \(x=V^\pi(s)\).
All conditional identities below are understood almost surely over such states.
Because the reward is terminal, \(\mathbb E[R_i\mid s]=x\).
The other trajectories remain independent of \(\tau_i\) after conditioning on \(s\), so \(K_{-i}\mid s\sim\operatorname{Binomial}(n-1,p)\) and \(K_{-i}\) is conditionally independent of \(R_i\).
Writing \(\overline R_{-i}=K_{-i}/(n-1)\), we obtain
\begin{equation}
 Y_i=R_i-\frac{R_i+K_{-i}}n
 =\frac{n-1}{n}R_i-\frac{K_{-i}}n
 =\frac{n-1}{n}(R_i-\overline R_{-i}).
\end{equation}
It follows that
\begin{equation}
 \mathbb E[Y_i\mid s]
 =\frac{n-1}{n}\mathbb E[R_i\mid s]-\frac1n\mathbb E[K_{-i}\mid s]
 =\frac{n-1}{n}(x-p).
\end{equation}
Conditional independence makes \(\operatorname{Cov}(R_i,K_{-i}\mid s)=0\), hence
\begin{equation}
\begin{aligned}
 \operatorname{Var}(Y_i\mid s)
 &=\frac{(n-1)^2}{n^2}\operatorname{Var}(R_i\mid s)
   +\frac1{n^2}\operatorname{Var}(K_{-i}\mid s)
   -\frac{2(n-1)}{n^2}\operatorname{Cov}(R_i,K_{-i}\mid s)\\
 &=\frac{(n-1)^2}{n^2}x(1-x)+\frac{n-1}{n^2}p(1-p).
\end{aligned}
\label{eq:app-unfiltered-target}
\end{equation}
For a scalar prediction \(f\), write \(m(s)=\mathbb E[Y_i\mid s]\).
Since \(\mathbb E[m(s)-Y_i\mid s]=0\), the conditional squared loss decomposes as
\begin{equation}
\begin{aligned}
 \mathbb E[(f-Y_i)^2\mid s]
 &=\mathbb E[((f-m(s))+(m(s)-Y_i))^2\mid s]\\
 &=(f-m(s))^2
   +2(f-m(s))\mathbb E[m(s)-Y_i\mid s]
   +\mathbb E[(Y_i-m(s))^2\mid s]\\
 &=(f-m(s))^2+\operatorname{Var}(Y_i\mid s)\\
 &=\left[f-\frac{n-1}{n}(x-p)\right]^2
   +\frac{(n-1)^2}{n^2}x(1-x)+\frac{n-1}{n^2}p(1-p).
\end{aligned}
\label{eq:app-residual-projection}
\end{equation}

\paragraph{Conditioning on mixed groups.}
Let \(\mathcal M=\{0<K<n\}\).
On \(\{R_i=1\}\), this event requires \(K_{-i}<n-1\), whereas on \(\{R_i=0\}\) it requires \(K_{-i}>0\).
Conditional independence therefore gives the joint probability
\begin{equation}
 \Pr(R_i=1,\mathcal M\mid s)
 =\Pr(R_i=1\mid s)\Pr(K_{-i}<n-1\mid s)
 =x\bigl(1-p^{n-1}\bigr).
\end{equation}
For the unsuccessful outcome, it gives
\begin{equation}
 \Pr(R_i=0,\mathcal M\mid s)
 =\Pr(R_i=0\mid s)\Pr(K_{-i}>0\mid s)
 =(1-x)\bigl(1-(1-p)^{n-1}\bigr).
\end{equation}
These identities also hold at \(x=0\) and \(x=1\).
Adding the disjoint events yields
\begin{equation}
\begin{aligned}
 \Pr(\mathcal M\mid s)
 &=\Pr(R_i=1,\mathcal M\mid s)+\Pr(R_i=0,\mathcal M\mid s)\\
 &=[1-p^{n-1}]x+[1-(1-p)^{n-1}](1-x)\\
 &=1-xp^{n-1}-(1-x)(1-p)^{n-1}>0.
\end{aligned}
\label{eq:app-mixed-probability}
\end{equation}
Positivity holds for every \(x\in[0,1]\) because both \(1-p^{n-1}\) and \(1-(1-p)^{n-1}\) are positive.
On \(\mathcal M^c\), all rewards coincide, so \(Y_i=0\).
Consequently,
\begin{equation}
 \mathbb E[Y_i\mathbf1_{\mathcal M}\mid s]
 =\mathbb E[Y_i\mid s]-\mathbb E[Y_i\mathbf1_{\mathcal M^c}\mid s]
 =\frac{n-1}{n}(x-p).
\end{equation}
The conditional mean on mixed groups is therefore
\begin{equation}
\begin{aligned}
 F^\star(s)
 &=\mathbb E[Y_i\mid s,\mathcal M]\\
 &=\frac{\mathbb E[Y_i\mathbf1_{\mathcal M}\mid s]}{\Pr(\mathcal M\mid s)}\\
 &=\frac{(n-1)(x-p)}
         {n\bigl[1-xp^{n-1}-(1-x)(1-p)^{n-1}\bigr]}.
\end{aligned}
\label{eq:app-mixed-target-proof}
\end{equation}
Using \(\mathbb E[F^\star(s)-Y_i\mid s,\mathcal M]=0\), the corresponding loss is
\begin{equation}
\begin{aligned}
 \mathbb E[(f-Y_i)^2\mid s,\mathcal M]
 &=(f-F^\star(s))^2
   +2(f-F^\star(s))\mathbb E[F^\star(s)-Y_i\mid s,\mathcal M]\\
 &\quad+\mathbb E[(F^\star(s)-Y_i)^2\mid s,\mathcal M]\\
 &=(f-F^\star(s))^2+\operatorname{Var}(Y_i\mid s,\mathcal M)\\
 &=\left[
       f-\frac{(n-1)(x-p)}
         {n\bigl[1-xp^{n-1}-(1-x)(1-p)^{n-1}\bigr]}
    \right]^2
    +\operatorname{Var}(Y_i\mid s,\mathcal M).
\end{aligned}
\end{equation}
The variance term does not depend on \(f\), so the unique minimizer over scalar predictions is \(f=F^\star(s)\).
This identifies the population target without assuming that a parameterized critic attains it.

\paragraph{Monotonicity and value recovery.}
Differentiating the explicit fraction yields
\begin{equation}
\begin{aligned}
 &\frac{d}{dx}\left[
   \frac{(n-1)(x-p)}
        {n\bigl[1-xp^{n-1}-(1-x)(1-p)^{n-1}\bigr]}
   \right]\\
 &\quad=\frac{n-1}{n}\,
   \frac{
     1-xp^{n-1}-(1-x)(1-p)^{n-1}
     -(x-p)\bigl[(1-p)^{n-1}-p^{n-1}\bigr]
   }{\bigl[1-xp^{n-1}-(1-x)(1-p)^{n-1}\bigr]^2}\\
 &\quad=\frac{n-1}{n}\,
   \frac{1-(1-p)^{n-1}+p\bigl[(1-p)^{n-1}-p^{n-1}\bigr]}
        {\bigl[1-xp^{n-1}-(1-x)(1-p)^{n-1}\bigr]^2}\\
 &\quad=\frac{(n-1)\bigl[1-p^n-(1-p)^n\bigr]}
        {n\bigl[1-xp^{n-1}-(1-x)(1-p)^{n-1}\bigr]^2}>0.
\end{aligned}
\label{eq:app-mixed-target-monotonicity}
\end{equation}
Indeed, \(p^n<p\) and \((1-p)^n<1-p\) for \(n\ge2\), so the numerator is positive.
The denominator is nonzero throughout \([0,1]\).
The mean value theorem establishes strict ordering in \(x\), including the endpoints, and completes the proof of Eq.~\eqref{eq:theory-residual-target}.

For fixed known \(p\), set \(u=F^\star(s)\).
Clearing the denominator gives
\begin{equation}
 n u\bigl[1-xp^{n-1}-(1-x)(1-p)^{n-1}\bigr]=(n-1)(x-p).
\end{equation}
Expanding the terms in \(x\) yields
\begin{equation}
 n\bigl[1-(1-p)^{n-1}\bigr]u
 +n\bigl[(1-p)^{n-1}-p^{n-1}\bigr]u x
 =(n-1)x-(n-1)p.
\end{equation}
Collecting them on the left gives
\begin{equation}
 \left[n-1-n\bigl((1-p)^{n-1}-p^{n-1}\bigr)u\right]x
 =n\bigl[1-(1-p)^{n-1}\bigr]u+(n-1)p.
\end{equation}
For every attainable \(u=F^\star(s)\), its coefficient is strictly positive:
\begin{equation}
\begin{aligned}
 &n-1-n\bigl[(1-p)^{n-1}-p^{n-1}\bigr]F^\star(s)\\
 &\quad=(n-1)\left[
       1-\frac{\bigl[(1-p)^{n-1}-p^{n-1}\bigr](x-p)}
                    {1-xp^{n-1}-(1-x)(1-p)^{n-1}}
       \right]\\
 &\quad=\frac{(n-1)\bigl[1-p^n-(1-p)^n\bigr]}
                    {1-xp^{n-1}-(1-x)(1-p)^{n-1}}>0.
\end{aligned}
\end{equation}
Division therefore recovers the value as
\begin{equation}
 V^\pi(s)
 =\frac{n\bigl[1-(1-p)^{n-1}\bigr]u+(n-1)p}
        {n-1-n\bigl[(1-p)^{n-1}-p^{n-1}\bigr]u}.
\label{eq:app-residual-inverse}
\end{equation}

\subsection{Proof of Proposition~\ref{prop:single-reconstruction}}
\label{app:single-proof}
\paragraph{Outcome martingale and TD residuals.}
Define \(M_t=\mathbb E_\pi[R\mid\mathcal F_t]\) and \(D_t=M_{t+1}-M_t\).
Because \(R_{<t}\) is \(\mathcal F_t\)-measurable,
\begin{equation}
 M_t=\mathbb E_\pi[R_{<t}+G_t\mid\mathcal F_t]
    =R_{<t}+\mathbb E_\pi[G_t\mid\mathcal F_t]
    =R_{<t}+V_t^\pi.
\end{equation}
The initial history contains only the fixed prompt, so \(M_0=p\).
At the terminal horizon, the reward has been observed and \(V_H^\pi=0\), giving \(M_H=R\).
The tower property shows that \((M_t)\) is a martingale:
\begin{equation}
 \mathbb E_\pi[M_{t+1}\mid\mathcal F_t]
 =\mathbb E_\pi[\mathbb E_\pi[R\mid\mathcal F_{t+1}]\mid\mathcal F_t]
 =\mathbb E_\pi[R\mid\mathcal F_t]=M_t.
\end{equation}
Since \(R_{<t+1}-R_{<t}=r_t\), its increments satisfy
\begin{equation}
 D_t=(R_{<t+1}+V_{t+1}^\pi)-(R_{<t}+V_t^\pi)
    =r_t+V_{t+1}^\pi-V_t^\pi=\delta_t^\pi.
\end{equation}
Their conditional means vanish:
\begin{equation}
 \mathbb E_\pi[D_t\mid\mathcal F_t]
 =\mathbb E_\pi[M_{t+1}\mid\mathcal F_t]-M_t=0.
\end{equation}
Telescoping over the fixed horizon gives
\begin{equation}
 R-p=M_H-M_0=\sum_{t=0}^{H-1}D_t.
\label{eq:app-outcome-martingale}
\end{equation}
All increments and policy scores after termination are zero, so the same identities apply to trajectories of different lengths.

\paragraph{Cancellation of past and future increments.}
Substituting Eq.~\eqref{eq:app-outcome-martingale} into the score identity yields
\begin{equation}
\begin{aligned}
 \nabla_\theta p
 &=\mathbb E_\pi\!\left[\left(\sum_{t=0}^{H-1}\psi_t\right)
                                  \left(\sum_{k=0}^{H-1}D_k\right)\right]\\
 &=\sum_{t=0}^{H-1}\sum_{k=0}^{H-1}\mathbb E_\pi[\psi_tD_k]\\
 &=\sum_t\sum_{k<t}\mathbb E_\pi[\psi_tD_k]
       +\sum_t\mathbb E_\pi[\psi_tD_t]
       +\sum_t\sum_{k>t}\mathbb E_\pi[\psi_tD_k].
\end{aligned}
\end{equation}
If \(k<t\), then \(D_k\) is \(\mathcal F_{k+1}\)-measurable and \(\mathcal F_{k+1}\subseteq\mathcal F_t\).
Conditioning on \(\mathcal F_t\) eliminates the policy score:
\begin{equation}
 \mathbb E_\pi[\psi_tD_k]
 =\mathbb E_\pi[\mathbb E_\pi[\psi_tD_k\mid\mathcal F_t]]
 =\mathbb E_\pi[D_k\mathbb E_\pi[\psi_t\mid\mathcal F_t]]=0.
\end{equation}
If \(k>t\), then \(\psi_t\) is \(\mathcal F_{t+1}\)-measurable and \(\mathcal F_{t+1}\subseteq\mathcal F_k\).
Conditioning on \(\mathcal F_k\) instead eliminates the future increment:
\begin{equation}
 \mathbb E_\pi[\psi_tD_k]
 =\mathbb E_\pi[\mathbb E_\pi[\psi_tD_k\mid\mathcal F_k]]
 =\mathbb E_\pi[\psi_t\mathbb E_\pi[D_k\mid\mathcal F_k]]=0.
\end{equation}
Only the matching indices remain, proving the first identity:
\begin{equation}
 \nabla_\theta p
 =\sum_t\mathbb E_\pi[\psi_tD_t]
 =\mathbb E_\pi\!\left[\sum_t\psi_t\delta_t^\pi\right].
\label{eq:app-diagonal-gradient}
\end{equation}

\paragraph{Generalized advantage estimation.}
For fixed \(0<\lambda\le1\), unrolling the GAE recursion with \(\widehat A_H^\lambda=0\) gives
\begin{equation}
\begin{aligned}
 \widehat A_t^\lambda(V^\pi)
 &=\delta_t^\pi+\lambda\widehat A_{t+1}^\lambda(V^\pi)\\
 &=\delta_t^\pi+\lambda\delta_{t+1}^\pi+\lambda^2\widehat A_{t+2}^\lambda(V^\pi)\\
 &=\sum_{k=t}^{H-1}\lambda^{k-t}\delta_k^\pi.
\end{aligned}
\end{equation}
The second line applies when \(t\le H-2\), and the last decision has \(\widehat A_{H-1}^\lambda=\delta_{H-1}^\pi\).
Since \(\lambda\) is deterministic, the cancellation above yields
\begin{equation}
\begin{aligned}
 \mathbb E_\pi[\psi_t\widehat A_t^\lambda(V^\pi)]
 &=\sum_{k=t}^{H-1}\lambda^{k-t}\mathbb E_\pi[\psi_tD_k]\\
 &=\mathbb E_\pi[\psi_tD_t]
       +\sum_{k=t+1}^{H-1}\lambda^{k-t}
          \mathbb E_\pi[\psi_t\mathbb E_\pi[D_k\mid\mathcal F_k]]\\
 &=\mathbb E_\pi[\psi_t\delta_t^\pi].
\end{aligned}
\end{equation}
Summing over decisions proves
\begin{equation}
 \mathbb E_\pi\!\left[\sum_t\psi_t\widehat A_t^\lambda(V^\pi)\right]
 =\sum_t\mathbb E_\pi[\psi_t\delta_t^\pi]
 =\nabla_\theta p.
\label{eq:app-gae-gradient-proof}
\end{equation}
For \(\lambda=0\), the recursion gives \(\widehat A_t^0(V^\pi)=\delta_t^\pi\), so Eq.~\eqref{eq:app-diagonal-gradient} covers this endpoint directly.
Together with Proposition~\ref{prop:group-identification}, the result gives
\begin{equation}
 \mathbb E\widehat g_{\mathrm{grp}}
 =\omega_n(p)\nabla_\theta p
 =\omega_n(p)\mathbb E_\pi\!\left[\sum_t\psi_t\widehat A_t^\lambda(V^\pi)\right].
\label{eq:app-group-single-bridge}
\end{equation}
This identity relates expected gradients and requires exact values for the single-trajectory estimator.
It does not assert equality of individual samples or of their variances.

\subsection{Behavior policies and importance weighting}
\label{app:offpolicy}
Let \(\beta_t(\cdot\mid\mathcal F_t)\) be the behavior policy used to sample action \(a_t\).
Any policy-version information needed to specify this conditional law is included in the history.
The target and behavior laws share the initial prompt and the transition and reward kernels.
Assume the behavior policy assigns positive probability to every action with positive target probability at the histories under consideration.
Behavior probabilities are held fixed when differentiating the target policy.
Define
\begin{equation}
 \rho_t=\frac{\pi_\theta(a_t\mid s_t)}{\beta_t(a_t\mid\mathcal F_t)}.
\end{equation}
Write \(W_{0:t}=\prod_{j=0}^t\rho_j\) for the prefix importance weight.
Products \(\rho_t\psi_t\) at zero target probability are interpreted as \(\nabla_\theta\pi_\theta(a_t\mid s_t)/\beta_t(a_t\mid\mathcal F_t)\).

\paragraph{What a single action ratio corrects.}
For any integrable function \(b(\mathcal F_t,a_t)\), conditioning on the sampled action gives
\begin{equation}
\mathbb E_\beta[\rho_t b(\mathcal F_t,a_t)\mid\mathcal F_t]
 =\sum_a\beta_t(a\mid\mathcal F_t)
       \frac{\pi_\theta(a\mid s_t)}{\beta_t(a\mid\mathcal F_t)}b(\mathcal F_t,a)
 =\sum_a\pi_\theta(a\mid s_t)b(\mathcal F_t,a).
\end{equation}
Thus the ratio corrects the conditional action distribution.
To distinguish this from correcting the history distribution, define
\begin{equation}
 Q_t^\pi(s,a)=\mathbb E[r_t+V_{t+1}^\pi\mid s_t=s,a_t=a].
\end{equation}
The expectation uses the common transition and reward kernels.
With \(\psi_t(s,a)=\nabla_\theta\log\pi_\theta(a\mid s)\), let
\begin{equation}
 f_t(s)=\sum_a\pi_\theta(a\mid s)\psi_t(s,a)
                  \bigl(Q_t^\pi(s,a)-V_t^\pi(s)\bigr).
\end{equation}
A second conditioning step gives
\begin{equation}
\begin{aligned}
 \mathbb E_\beta[\rho_t\psi_t\delta_t^\pi\mid\mathcal F_t]
 &=\sum_a\pi_\theta(a\mid s_t)\psi_t(s_t,a)
       \mathbb E[\delta_t^\pi\mid s_t,a_t=a]\\
 &=\sum_a\pi_\theta(a\mid s_t)\psi_t(s_t,a)
       \bigl(Q_t^\pi(s_t,a)-V_t^\pi(s_t)\bigr)=f_t(s_t).
\end{aligned}
\end{equation}
The same conditional expectation under \(\pi\), without a ratio, is also \(f_t(s_t)\).
Writing \(d_t^\beta\) and \(d_t^\pi\) for the two state distributions therefore yields
\begin{equation}
\mathbb E_\beta[\rho_t\psi_t\delta_t^\pi]
 -\mathbb E_\pi[\psi_t\delta_t^\pi]
 =\int f_t(s)\,d_t^\beta(ds)-\int f_t(s)\,d_t^\pi(ds)
 =\int f_t(s)(d_t^\beta-d_t^\pi)(ds).
\label{eq:app-token-is-error}
\end{equation}

\paragraph{Prefix correction and future TD residuals.}
On the history through \(a_t\), \(r_t\), and \(s_{t+1}\), cancellation of the common environment kernels gives
\begin{equation}
 \frac{d(P_\pi|_{\mathcal F_{t+1}})}{d(P_\beta|_{\mathcal F_{t+1}})}
 =\prod_{j=0}^t
      \frac{\pi_\theta(a_j\mid s_j)}{\beta_j(a_j\mid\mathcal F_j)}
 =W_{0:t}.
\end{equation}
Since \(\psi_t\delta_t^\pi\) is \(\mathcal F_{t+1}\)-measurable, change of measure yields
\begin{equation}
 \mathbb E_\beta\!\left[\sum_tW_{0:t}\psi_t\delta_t^\pi\right]
 =\sum_t\mathbb E_\pi[\psi_t\delta_t^\pi]
 =\nabla_\theta p.
\label{eq:app-offpolicy-gradient}
\end{equation}
This is a reference identity for prefix weights.
The implemented action ratio \(\rho_t\) alone does not supply these weights.

Future TD residuals introduce another discrepancy for GAE.
The Bellman identity \(V_k^\pi(s)=\sum_a\pi_\theta(a\mid s)Q_k^\pi(s,a)\) gives
\begin{equation}
\begin{aligned}
 m_k^\beta
 :=\mathbb E_\beta[\delta_k^\pi\mid\mathcal F_k]
 &=\sum_a\beta_k(a\mid\mathcal F_k)
       \bigl(Q_k^\pi(s_k,a)-V_k^\pi(s_k)\bigr)\\
 &=\sum_a\beta_k(a\mid\mathcal F_k)Q_k^\pi(s_k,a)-V_k^\pi(s_k)\\
 &=\sum_a\bigl(\beta_k(a\mid\mathcal F_k)-\pi_\theta(a\mid s_k)\bigr)Q_k^\pi(s_k,a).
\end{aligned}
\end{equation}
For \(k>t\), the factor \(\rho_t\psi_t\) is \(\mathcal F_k\)-measurable, so
\begin{equation}
 \mathbb E_\beta[\rho_t\psi_t\delta_k^\pi]
 =\mathbb E_\beta[\rho_t\psi_t\mathbb E_\beta[\delta_k^\pi\mid\mathcal F_k]]
 =\mathbb E_\beta[\rho_t\psi_t m_k^\beta].
\end{equation}
Expanding GAE and applying Eq.~\eqref{eq:app-token-is-error} produces
\begin{equation}
\begin{aligned}
 &\mathbb E_\beta\!\left[\sum_t\rho_t\psi_t\widehat A_t^\lambda(V^\pi)\right]-\nabla_\theta p\\
 &\quad=\sum_t\left(\mathbb E_\beta[\rho_t\psi_t\delta_t^\pi]
                     -\mathbb E_\pi[\psi_t\delta_t^\pi]\right)
       +\sum_{t<k}\lambda^{k-t}\mathbb E_\beta[\rho_t\psi_t\delta_k^\pi]\\
 &\quad=\sum_t\int f_t(s)(d_t^\beta-d_t^\pi)(ds)
       +\sum_{t<k}\lambda^{k-t}\mathbb E_\beta[\rho_t\psi_t m_k^\beta].
\end{aligned}
\label{eq:app-behavior-gae-error}
\end{equation}
The two terms describe differences in the visited states and in subsequent actions, respectively.
Both vanish under on-policy sampling.

\subsection{Calibration and the implemented update}
\label{app:theory-boundaries}
\paragraph{What residual supervision identifies.}
For two real states of the same prompt, Proposition~\ref{prop:group-identification} implies
\begin{equation}
 \operatorname{sign}(F^\star(s')-F^\star(s))
 =\operatorname{sign}(V^\pi(s')-V^\pi(s)).
\end{equation}
At the root, \(F^\star(s_0)=0\) for every \(p\), so this prediction alone does not identify the prompt success probability.
The transformation in Eq.~\eqref{eq:theory-residual-target} can be nonlinear and need not preserve conditional expectations across stochastic transitions.
Thus ordering alone does not justify substituting \(F^\star\) for \(V^\pi\) in Proposition~\ref{prop:single-reconstruction}.

For a nonnegative divergence mask or sample weight \(c\), let \(\mathcal L_c(f\mid s)=\mathbb E[c\mathbf1_{\mathcal M}(f-Y_i)^2\mid s]\).
Assume its weighted second moment is finite and \(0<\mathbb E[c\mathbf1_{\mathcal M}\mid s]<\infty\).
Expanding the quadratic gives
\begin{equation}
 \mathcal L_c(f\mid s)
 =f^2\mathbb E[c\mathbf1_{\mathcal M}\mid s]
  -2f\mathbb E[c\mathbf1_{\mathcal M}Y_i\mid s]
  +\mathbb E[c\mathbf1_{\mathcal M}Y_i^2\mid s].
\end{equation}
Its leading coefficient is positive, so setting its derivative to zero yields the unique minimizer
\begin{equation}
 F_c^\star(s)
 =\frac{\mathbb E[c\mathbf1_{\mathcal M}Y_i\mid s]}
         {\mathbb E[c\mathbf1_{\mathcal M}\mid s]}
 =\frac{\mathbb E[cY_i\mid s,\mathcal M]}{\mathbb E[c\mid s,\mathcal M]}.
\end{equation}
The change from the unweighted target is explicit:
\begin{equation}
F_c^\star(s)-F^\star(s)
 =\frac{\mathbb E[cY_i\mid s,\mathcal M]
         -\mathbb E[c\mid s,\mathcal M]\mathbb E[Y_i\mid s,\mathcal M]}
        {\mathbb E[c\mid s,\mathcal M]}
 =\frac{\operatorname{Cov}(c,Y_i\mid s,\mathcal M)}{\mathbb E[c\mid s,\mathcal M]}.
\end{equation}
Positive weights depending only on the state cancel, whereas masks depending on outcomes can change the target.
Random loss denominators, value clipping, and ranking regularization can further change the objective.
After switching, the base absolute-return loss instead satisfies
\begin{equation}
 \mathbb E_\pi[(f-G_t)^2\mid s_t=s]
 =(f-V^\pi(s))^2+\operatorname{Var}_\pi(G_t\mid s_t=s).
\end{equation}
This is the calibration required by Proposition~\ref{prop:single-reconstruction}.
Residual explained variance and ranking accuracy do not by themselves certify that the learned critic has attained it.

\paragraph{Approximate critic and advantage bias.}
Fix the critic during the actor update and define \(e_t=V_\phi(s_t)-V_t^\pi\), with zero error at terminal and absorbing states.
Write \(\delta_t^\phi=r_t+V_\phi(s_{t+1})-V_\phi(s_t)\).
Then \(\delta_t^\phi-\delta_t^\pi=e_{t+1}-e_t\).
For \(0<\lambda\le1\), shifting the summation index gives
\begin{equation}
\begin{aligned}
 \widehat A_t^\lambda(V_\phi)-\widehat A_t^\lambda(V^\pi)
 &=\sum_{k=t}^{H-1}\lambda^{k-t}(e_{k+1}-e_k)\\
 &=\sum_{j=t+1}^{H}\lambda^{j-t-1}e_j
     -\sum_{j=t}^{H-1}\lambda^{j-t}e_j\\
 &=-e_t+\lambda^{H-t-1}e_H
     +\sum_{j=t+1}^{H-1}\bigl(\lambda^{j-t-1}-\lambda^{j-t}\bigr)e_j\\
 &=-e_t+(1-\lambda)\sum_{j=t+1}^{H-1}\lambda^{j-t-1}e_j.
\end{aligned}
\end{equation}
The state baseline \(e_t\) contributes zero expected policy score.
Using Proposition~\ref{prop:single-reconstruction},
\begin{equation}
\begin{aligned}
 &\mathbb E_\pi\!\left[\sum_t\psi_t\widehat A_t^\lambda(V_\phi)\right]-\nabla_\theta p\\
 &\quad=\sum_t\mathbb E_\pi\!\left[
           \psi_t\bigl(\widehat A_t^\lambda(V_\phi)-\widehat A_t^\lambda(V^\pi)\bigr)\right]\\
 &\quad=-\sum_t\mathbb E_\pi[\psi_t e_t]
       +(1-\lambda)\sum_{t<k}\lambda^{k-t-1}\mathbb E_\pi[\psi_t e_k]\\
 &\quad=(1-\lambda)\sum_{t<k}\lambda^{k-t-1}\mathbb E_\pi[\psi_t e_k].
\end{aligned}
\label{eq:app-approximate-critic}
\end{equation}
For \(\lambda=1\), the critic error is only a baseline and the bias vanishes.
For \(0<\lambda<1\), future value errors enter through their correlation with earlier policy scores.
At \(\lambda=0\), the advantage error is directly \(e_{t+1}-e_t\), giving bias \(\sum_t\mathbb E_\pi[\psi_t e_{t+1}]\).

Under behavior sampling, the action ratio still eliminates the current baseline:
\begin{equation}
 \mathbb E_\beta[\rho_t\psi_t e_t]
 =\mathbb E_\beta\!\left[e_t\sum_a\pi_\theta(a\mid s_t)\psi_t(s_t,a)\right]=0.
\end{equation}
The same pathwise expansion therefore gives
\begin{equation}
 \mathbb E_\beta\!\left[\sum_t\rho_t\psi_t
       \bigl(\widehat A_t^\lambda(V_\phi)-\widehat A_t^\lambda(V^\pi)\bigr)\right]
 =(1-\lambda)\sum_{t<k}\lambda^{k-t-1}\mathbb E_\beta[\rho_t\psi_t e_k].
\label{eq:app-behavior-critic-error}
\end{equation}
These identities concern a fixed critic evaluated on fresh trajectories.
If the critic is fitted on the same trajectories, conditioning only on the decision history need not make its fitted prediction a valid baseline.

\paragraph{Masking, normalization, and adaptive traces.}
For the actor surrogate with stopped advantages, let \(c_t\in[0,1]\) be the active multiplier from DIS and policy clipping, away from their nondifferentiable boundaries.
To compare with a fixed normalization, set \(\bar g_q=H^{-1}\nabla_\theta p\) and define
\begin{equation}
 X_0=\frac1H\sum_{t<L}\rho_t\psi_t\widehat A_t^\lambda(V_\phi).
\end{equation}
The corresponding score contribution with trajectory length normalization and the realized trace parameter \(\lambda_L\) is
\begin{equation}
 X_{\mathrm{imp}}=\frac1L\sum_{t<L}c_t\rho_t\psi_t\widehat A_t^{\lambda_L}(V_\phi).
\end{equation}
Insert the intermediate updates \(H^{-1}\sum_{t<L}c_t\rho_t\psi_t\widehat A_t^\lambda(V_\phi)\) and \(H^{-1}\sum_{t<L}c_t\rho_t\psi_t\widehat A_t^{\lambda_L}(V_\phi)\).
Subtracting them and expanding the difference of the two GAE traces yields
\begin{equation}
\begin{aligned}
 X_{\mathrm{imp}}-X_0
 &=\frac1H\sum_{t<L}(c_t-1)\rho_t\psi_t\widehat A_t^\lambda(V_\phi)
   +\frac1H\sum_{t<L}c_t\rho_t\psi_t
       \sum_{k=t}^{L-1}(\lambda_L^{k-t}-\lambda^{k-t})\delta_k^\phi\\
 &\quad+\left(\frac1L-\frac1H\right)
       \sum_{t<L}c_t\rho_t\psi_t\widehat A_t^{\lambda_L}(V_\phi).
\end{aligned}
\end{equation}
By adding the behavior error in Eq.~\eqref{eq:app-behavior-gae-error}, the critic error in Eq.~\eqref{eq:app-behavior-critic-error}, and this pathwise difference,
\begin{equation}
\begin{aligned}
 \mathbb E_\beta X_{\mathrm{imp}}-\bar g_q
 &=\frac1H\sum_t\int f_t(s)(d_t^\beta-d_t^\pi)(ds)
   +\frac1H\sum_{t<k}\lambda^{k-t}\mathbb E_\beta[\rho_t\psi_t m_k^\beta]\\
 &\quad+\frac{1-\lambda}{H}\sum_{t<k}\lambda^{k-t-1}\mathbb E_\beta[\rho_t\psi_t e_k]
   +\mathbb E_\beta[X_{\mathrm{imp}}-X_0].
\end{aligned}
\label{eq:app-practical-decomposition}
\end{equation}
The formula separates behavior sampling, critic approximation, and the changes to the surrogate update.
Its expectations use the behavior trajectory law before any additional selection based on outcomes.
An exact critic alone does not make these other terms vanish.
In particular, \(\lambda_L\) and \(1/L\) can depend on future actions, so the conditional cancellations in Proposition~\ref{prop:single-reconstruction} do not automatically apply to them.

\subsection{Single-turn reasoning and agentic histories}
Between two model decisions, \(s_{t+1}\) includes any intervening tool observations.
The environment contributes no policy score, while its randomness remains in the TD residual.
Using the action value defined above, let \(A_t^\pi(s,a)=Q_t^\pi(s,a)-V_t^\pi(s)\) and \(\zeta_t=r_t+V_{t+1}^\pi-Q_t^\pi(s_t,a_t)\).
By definition,
\begin{equation}
 \mathbb E[\zeta_t\mid s_t,a_t]
 =\mathbb E[r_t+V_{t+1}^\pi\mid s_t,a_t]-Q_t^\pi(s_t,a_t)=0.
\end{equation}
The TD residual decomposes as
\begin{equation}
 \delta_t^\pi
 =Q_t^\pi(s_t,a_t)-V_t^\pi(s_t)+\zeta_t
 =A_t^\pi(s_t,a_t)+\zeta_t.
\end{equation}
Under the target policy, conditioning first on \((s_t,a_t)\) gives
\begin{equation}
\mathbb E_\pi[\psi_t\delta_t^\pi\mid s_t]
 =\mathbb E_\pi[\psi_t A_t^\pi(s_t,a_t)\mid s_t]
   +\mathbb E_\pi[\psi_t\mathbb E[\zeta_t\mid s_t,a_t]\mid s_t]
 =\mathbb E_\pi[\psi_t A_t^\pi(s_t,a_t)\mid s_t].
\end{equation}
For finite second moments, the law of total covariance yields
\begin{equation}
\begin{aligned}
 \operatorname{Cov}_\pi(\psi_t\delta_t^\pi\mid s_t)
 &=\operatorname{Cov}_\pi\!\left(\mathbb E[\psi_t\delta_t^\pi\mid s_t,a_t]\,\middle|\,s_t\right)
   +\mathbb E_\pi\!\left[\operatorname{Cov}(\psi_t\delta_t^\pi\mid s_t,a_t)\,\middle|\,s_t\right]\\
 &=\operatorname{Cov}_\pi(\psi_t A_t^\pi(s_t,a_t)\mid s_t)
   +\mathbb E_\pi\!\left[\psi_t\psi_t^\top
       \operatorname{Var}(r_t+V_{t+1}^\pi\mid s_t,a_t)\,\middle|\,s_t\right].
\end{aligned}
\end{equation}
Deterministic transitions and rewards make the second term zero.
Stochastic tool responses or rewards can make it nonzero, but tool use alone does not imply stochasticity.
Thus both task types satisfy the same policy-gradient identities, while the conditional variance and the difficulty of estimating values can differ.
These results do not establish critic convergence, a variance ordering between complete algorithms, or a higher asymptotic reward ceiling.

\endgroup
\twocolumn

\section{Related Work}
\label{sec:related}

\paragraph{Synchronous and Asynchronous RL.}
Standard Large Language Model (LLM) post-training collects a complete batch of rollouts under a fixed policy before each update \citep{ziegler-2019-preferences,stiennon-2020-summarize,ouyang-2022-instructgpt}.
Synchronous Reinforcement Learning (RL) systems~\citep{yao-2023-deepspeedchat,hu-2024-openrlhf,sheng-2025-hybridflow} implement this interleaved generation and training loop, which leaves accelerators idle while waiting for the longest rollout in a batch, especially when reasoning and agentic traces have long-tailed lengths.
Asynchronous RL systems~\citep{noukhovitch-2025-asynchronous,fu-2025-areal,piche-2025-pipelinerl} therefore decouple generation from learning and train as trajectories arrive, which improves training efficiency.
Recent LLM RL widely adopts Group Relative Policy Optimization (GRPO)~\citep{shao-2024-deepseekmath,deepseekai-2025-r1}, which estimates advantages by normalizing rewards within a prompt-level group.
However, this group-wise sampling does not fit asynchronous training, because an update must wait until every group member finishes and thus recreates a synchronization barrier with larger policy lag.
Recent methods stabilize optimization under this policy lag through off-policy objective design, variance control, or adaptive trust regions \citep{zheng2025prosperity,huang2026stableasync,yang2026sat}.
Single-rollout Asynchronous Optimization (SAO)~\citep{hou-2026-sao} instead replaces group-wise sampling with one rollout per prompt, uses token-level GAE from a learned critic for multi-turn agentic async RL, and stabilizes training with a scaled pretrained value model.
We reproduce SAO as \lisao in a lightweight setting without scaled value pretraining, initializing the value model from the policy checkpoint instead.

\paragraph{Group-Relative and Value-Based Optimization.}
Critic-free methods replace a learned value function with sampled baselines.
Beyond GRPO, RLOO uses leave-one-out baselines~\citep{ahmadian-etal-2024-back}, DAPO introduces dynamic sampling and asymmetric clipping~\citep{yu-2025-dapo}, and GSPO performs importance weighting and clipping at the sequence level~\citep{zheng-2025-gspo}.
Their reliance on group statistics reduces critic overhead but limits temporal credit assignment and delays updates until a complete group is available.
Value-based methods instead learn a critic and estimate advantages with Generalized Advantage Estimation~\citep{schulman-2017-ppo,schulman-2016-gae}.
VC-PPO and VAPO improve value learning through critic pretraining and decoupled GAE~\citep{yuan-2025-vcppo,yue-2025-vapo}, while SAPO shares an autoregressive backbone between policy and value prediction for single-rollout optimization~\citep{liang-2026-sapo}.
Unlike methods that remain group-relative or value-based throughout training, \liisa uses group-relative actor-critic bootstrapping, then switches to single-rollout GAE once the critic is calibrated.

\section{Experimental Details}
\label{app:exp-details}

\subsection{Experimental Configuration}
\label{app:hardware}
The experiments are conducted on a Linux server equipped with two Intel Xeon Platinum 8358 CPUs at 2.60\,GHz, 2\,TB RAM, and 8 NVIDIA A800-SXM4-80GB GPUs.
Our implementation is based on PyTorch~\citep{paszke-2019-pytorch} version 2.11.0, with CUDA version 12.9 and Python 3.12.3.
Rollouts are generated with SGLang~\citep{zheng-2024-sglang} version 0.5.15, and model training uses Megatron-LM~\citep{shoeybi-2019-megatron} version 0.16.0rc0.
We take this stack from the official \textsc{slime} Docker image\footnote{\url{https://hub.docker.com/r/slimerl/slime}} and replace the bundled \textsc{slime} code with our modified implementation.

\subsection{Datasets and Experimental Settings}
\label{app:data}

Training uses DAPO-Math-17K\footnote{\url{https://huggingface.co/datasets/zhuzilin/dapo-math-17k}}.
The single-turn setting starts from the Qwen3-4B checkpoint\footnote{\url{https://huggingface.co/Qwen/Qwen3-4B}}, a post-trained model rather than the pretrained base Qwen3-4B-Base, and generates a text solution.
The ReTool setting starts from a tool-use supervised checkpoint of Qwen3-4B-Instruct-2507\footnote{\url{https://huggingface.co/font-info/qwen3-4b-sft-SGLang-RL}} and interleaves reasoning with a Python interpreter on the same training problems.
Evaluation uses AIME 2024\footnote{\url{https://huggingface.co/datasets/zhuzilin/aime-2024}}, AIME 2025\footnote{\url{https://huggingface.co/datasets/zhuzilin/aime-2025}}, and AIME 2026\footnote{\url{https://huggingface.co/datasets/math-ai/aime26}}.
We draw 16 samples per problem at temperature $1.0$ and report the unbiased pass@$k$ estimator of \citet{chen-2021-codex}.
Single-turn models are evaluated without tools.
ReTool models are evaluated in the same interpreter sandbox used for training.
These task-specific differences are summarized in Table~\ref{tab:task-setup}.

\begin{table}[t]
    \centering
    \small
    \caption{Task-specific settings.}
    \label{tab:task-setup}
    \begin{tabular*}{0.95\columnwidth}{@{\hspace{\tabcolsep}\extracolsep{\fill}}lcc@{\hspace{\tabcolsep}}}
        \toprule
        \textbf{Setting} & \textbf{Single-turn} & \textbf{ReTool} \\
        \midrule
        Base Model & Qwen3-4B & tool-use SFT \\
        Generation & text, no tools & Python interpreter \\
        Max turns & 1 & 16 \\
        Timeout & none & 120\,s \\
        Evaluation & no tools & same sandbox \\
        \bottomrule
    \end{tabular*}
\end{table}

\subsection{Training and Evaluation Details}
\label{app:hparams}

Both algorithms share the configuration in Table~\ref{tab:shared-setup}.
Algorithm-specific choices are listed in Table~\ref{tab:algo-setup}.
We do not evaluate during training.
Checkpoints are saved every 100 steps and scored offline with the same reward used in training.
The outcome reward is $1$ when the boxed answer, or otherwise the last integer in the response, is mathematically equivalent to the label, and $0$ otherwise.
Responses that enter the final quarter of the context window additionally receive the overlong penalty of DAPO~\citep{yu-2025-dapo}.
GRPO uses the asymmetric clip of DAPO, while \lisao uses the token clip and DIS interval of SAO~\citep{hou-2026-sao}.

\begin{table}[t]
    \centering
    \small
    \caption{Shared training and evaluation configuration.}
    \label{tab:shared-setup}
    \begin{tabular*}{0.95\columnwidth}{@{\hspace{\tabcolsep}\extracolsep{\fill}}lc@{\hspace{\tabcolsep}}}
        \toprule
        \textbf{Setting} & \textbf{Value} \\
        \midrule
        Policy & Qwen3-4B, thinking off \\
        Training set & DAPO-Math-17K \\
        Context length & 32{,}768 \\
        Reward & binary outcome \\
        Overlong penalty & last quarter, factor $1$ \\
        Rollout steps & 500 \\
        Global batch size & 128 \\
        Max policy lag & 16 \\
        Optimizer & Adam, constant schedule \\
        Learning rate & $1\times 10^{-6}$ \\
        Weight decay & $0.1$ \\
        Adam betas & $(0.9,\ 0.98)$ \\
        Rollout temperature & $1.0$ \\
        Rollout top-$p$ & $1.0$ \\
        Evaluation & AIME 2024/2025/2026 \\
        Samples per problem & 16 \\
        Evaluation temperature & $1.0$ \\
        Evaluation top-$p$ & $1.0$ \\
        \bottomrule
    \end{tabular*}
\end{table}

\begin{table}[t]
    \centering
    \small
    \caption{Algorithm-specific configuration for the GRPO and \lisao comparison.}
    \label{tab:algo-setup}
    \begin{tabular*}{0.95\columnwidth}{@{\hspace{\tabcolsep}\extracolsep{\fill}}lcc@{\hspace{\tabcolsep}}}
        \toprule
        \textbf{Setting} & \textbf{GRPO} & \textbf{\lisao} \\
        \midrule
        Samples per prompt & 8 & 1 \\
        Prompts per batch & 16 & 128 \\
        Advantage & group-relative & GAE \\
        Clip range & $0.2$ / $0.28$ & $0.3$ / $5.0$ \\
        DIS interval & none & $[0.7,\ 6.0]$ \\
        Critic & none & frozen attention \\
        Critic learning rate & none & $5\times 10^{-6}$ \\
        Critic warmup & none & 10 steps \\
        Critic updates $K$ & none & 2 \\
        \bottomrule
    \end{tabular*}
\end{table}

\section{Reproduction Details of \lisao}
\label{app:lisao-repro}

This section records the choices required to run SAO without its scaled value model.
The comparison in Section~\ref{sec:lisao} uses this implementation.

\subsection{Critic Initialization}
SAO relies on a value model produced by scaled value pretraining, and neither that model nor its training pipeline is publicly available.
\lisao therefore keeps a second copy of the policy parameters and initializes it from the same checkpoint.
The critic output layer is replaced by a scalar value head.
An ablation in SAO shows that freezing multi-head attention and updating only the MLP blocks regularizes critic optimization, whereas full-parameter value training yields larger gradients and lower accuracy~\citep{hou-2026-sao}.
We follow this validated setting: parameters whose names contain \texttt{self\_attention} stay frozen, and training updates the MLP blocks together with the value head.
The actor is held fixed for the first 10 rollout steps while the critic is updated.
After that warmup, each batch applies one policy update and $K=2$ critic updates.
The critic learning rate is $5\times 10^{-6}$, matching the value-model rate reported for SAO.

\subsection{Advantage Estimation}
The policy advantage uses length-adaptive GAE with $\alpha=1.5$, so the per-trajectory coefficient is $\lambda=1-1/(\alpha L)$ and is clamped to $[0,1]$~\citep{yue-2025-vapo,hou-2026-sao}.
The discount is $\gamma=1$.
The critic target uses $\lambda=1$.
On multi-turn ReTool trajectories, GAE is evaluated only on model-generated tokens, which is the skip-observation recursion in Eq.(\ref{eq:skip_obs_td}) and Eq.(\ref{eq:skip_obs_gae}).
Single-turn responses consist only of model-generated tokens, so the same estimator reduces to ordinary GAE.
DIS reads the rollout log probability stored at sampling time and drops tokens whose importance ratio falls outside $[0.7, 6.0]$.
The entropy coefficient is $0$.

\begin{figure*}[t]
    \centering
    \includegraphics[width=\textwidth,height=0.78\textheight,keepaspectratio]{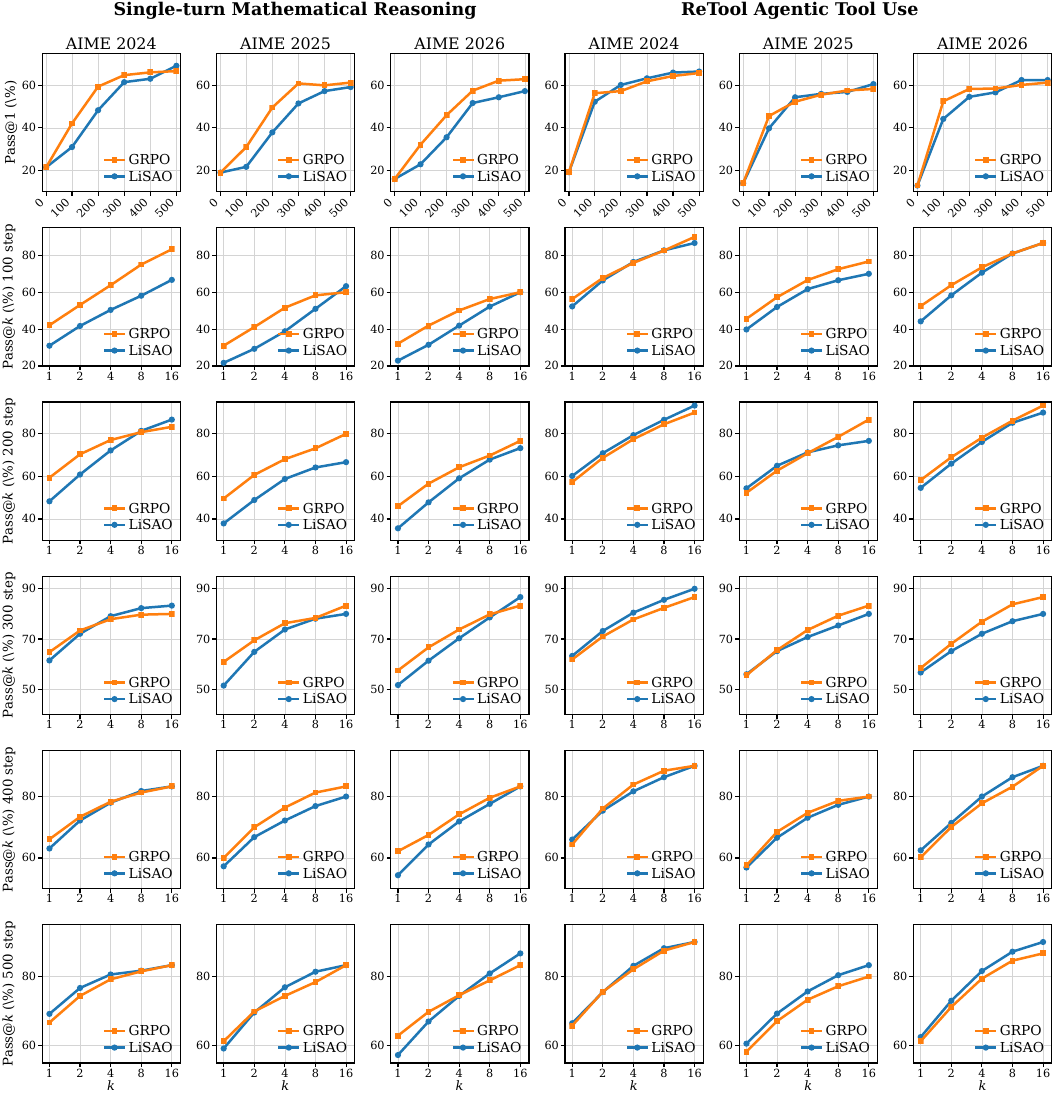}
    \caption{
    \textbf{GRPO vs. \lisao} on single-turn mathematical reasoning (left) and ReTool agentic tool use (right).
    Results for each task are reported separately on AIME 2024, 2025, and 2026.
    The top row is pass@1 from the shared initialization to step 500.
    Each lower row is pass@$k$ at the labeled checkpoint.
    }
    \label{fig:grpo_vs_lisao_full}
\end{figure*}

\subsection{Tasks and In-flight Updates}
Single-turn training starts from the post-trained Qwen3-4B checkpoint and generates text solutions.
ReTool training starts from a tool-use supervised checkpoint of Qwen3-4B and generates with a Python interpreter~\citep{feng-2025-retool}.
In both settings, an unfinished rollout keeps its generated prefix when the trainer publishes a new policy and continues from that prefix.
A single trajectory may therefore contain tokens sampled from more than one policy version.
The token-level ratio uses the log probability stored for the version that sampled the token.
Held-out AIME evaluation for the single-turn models is text-only.
The ReTool models are evaluated in the same interpreter sandbox used for training.

\section{Full Comparison of GRPO and \lisao}
\label{app:grpo-vs-lisao-full}

Figure~\ref{fig:grpo_vs_lisao_full} reports the same comparison separately for each year.
The left three columns are single-turn mathematical reasoning and the right three are ReTool agentic tool use.
The top row tracks pass@1 from the shared initialization to step 500.
Each lower row plots pass@$k$ at one checkpoint, with $k\in\{1,2,4,8,16\}$.

\section{\texorpdfstring{Details of \liisa}{Details of Li2SAO}}
\label{app:li2sao-alg}

\subsection{Training and Evaluation Configuration}

\liisa uses the training and evaluation configuration in Table~\ref{tab:shared-setup}.
The clip range, DIS interval, frozen-attention critic, critic learning rate, and \(K=2\) critic updates match \lisao in Table~\ref{tab:algo-setup}.
Length-adaptive GAE uses the coefficient in Appendix~\ref{app:lisao-repro}.
Table~\ref{tab:li2sao-setup} lists the remaining choices.
The trajectory budget stays at 128, so the group size \(n_b=8\) uses 16 prompts per batch and \(n_s=1\) uses 128.

\begin{table}[t]
    \centering
    \small
    \caption{\liisa training configuration. Shared training choices are in Table~\ref{tab:shared-setup}. 
    The clip range, DIS interval, critic architecture, critic learning rate, and \(K\) match \lisao in Table~\ref{tab:algo-setup}.
    }
    \label{tab:li2sao-setup}
    \begin{tabular*}{0.95\columnwidth}{@{\hspace{\tabcolsep}\extracolsep{\fill}}lcc@{\hspace{\tabcolsep}}}
        \toprule
        \textbf{Setting} & \textbf{Bootstrap} & \textbf{After switch} \\
        \midrule
        Samples per prompt & 8 & 1 \\
        Prompts per batch & 16 & 128 \\
        Advantage & group-relative & GAE \\
        Critic target & group-centered & absolute \\
        Value clip \(\epsilon_v\) & 0.2 & inactive \\
        Ranking \(\alpha\), \(m\) & 0.1, 0.1 & inactive \\
        \((\beta,\eta_e,\eta_a,M)\) & \multicolumn{2}{c}{\((0.9,\ 0.2,\ 0.65,\ 10)\)} \\
        \((T_{\min},T_{\max})\) & \multicolumn{2}{c}{\((60,\ 150)\)} \\
        \bottomrule
    \end{tabular*}
\end{table}

\begin{algorithm}[t]
    \caption{Algorithmic pseudocode of \liisa.}
    \label{alg:li2sao}
    \begin{algorithmic}
    \relpenalty=10000
    \binoppenalty=10000
    \REQUIRE Dataset \(\mathcal{D}\), policy \(\pi_\theta\), critic \(V_\phi\), budget \(B\), sizes \(n_b\) and \(n_s=1\), and \((\beta,\allowbreak\eta_e,\allowbreak\eta_a,\allowbreak M,\allowbreak T_{\min},\allowbreak T_{\max})\).
    \ENSURE Trained policy \(\pi_\theta\).
    \STATE Initialize \(V_\phi\) from \(\pi_\theta\); set \(n\gets n_b\), \(c\gets 0\), \(\mathrm{switched}\gets\mathrm{false}\)
    \FOR{\(k=1,2,\ldots\)}
        \STATE Collect \(B\) in-flight trajectories, \(n\) per prompt; drop over-lag samples
        \IF{\(\mathrm{switched}=\mathrm{false}\)}
            \STATE In each group \(g\), set \(\mu_g,\sigma_g\) and \(\hat A_{i,t}^{\mathrm{actor}}\gets\allowbreak(R_i-\mu_g)\allowbreak/(\sigma_g+\varepsilon)\) on model tokens
            \STATE Update \(\theta\) by DIS, Eq.(\ref{eq:dis_loss})
            \STATE On mixed groups, mask tokens before \(d_i\) and set \(\widetilde G_{i,t}\gets\allowbreak\hat G_{i,t}-\mu_g\)
            \STATE Update \(\phi\) by \(\mathcal{L}_{\mathrm{res}}+\alpha\mathcal{L}_{\mathrm{rank}}\), and update \(\bar e_k,\bar a_k\)
            \IF{\(k\ge T_{\min}\), \(\bar e_k>\eta_e\), and \(\bar a_k>\eta_a\)}
                \STATE \(c\gets c+1\)
            \ELSE
                \STATE \(c\gets 0\)
            \ENDIF
            \IF{\(c\ge M\) or \(k\ge T_{\max}\)}
                \STATE Finish the prefetched job; drop leftovers; \(n\gets n_s\); \(\mathrm{switched}\gets\mathrm{true}\)
            \ENDIF
        \ELSE
            \STATE \(\hat A_t\gets\) GAE on model tokens; update \(\theta\) by DIS; fit \(V_\phi\) to \(\hat G_t\)
        \ENDIF
    \ENDFOR
    \RETURN \(\pi_\theta\)
    \end{algorithmic}
    \end{algorithm}

\subsection{Algorithmic Pseudocode}

Algorithm~\ref{alg:li2sao} is the training loop of this procedure on \textsc{slime}.
The critic starts from the same checkpoint as the policy, and the two networks are updated from the first step.
Each step waits for a completed batch of 128 trajectories while the rollout workers continue generating under the current policy.
Trajectories whose policy lag exceeds the shared limit are dropped before the update.
During the bootstrap, those trajectories form groups of eight responses to each prompt.

\subsection{Execution on the In-Flight Pipeline}

\textsc{slime} builds generalized advantage estimates, whitens them across the data parallel group, and stores the corresponding returns before the optimizer step.
\liisa is inserted after that computation.
It overwrites actor advantages only when the largest prompt group in the collected batch has more than one response, using the raw outcome rewards of the complete group and writing the score onto tokens that enter the policy loss.
The stored returns are left in place.
Group centering and terminal ranking, when enabled, rewrite the critic targets in a later pass and do not alter the actor advantages.
If a data parallel shard cannot recover every member of a group, that shard keeps the advantage estimate already produced by the trainer.

Explained variance is reduced across data parallel workers before smoothing.
Each worker sends the count, sum, and sum of squares of its terminal outcomes and values, and the driver computes one variance from those totals.
The driver also owns the readiness counter and the outstanding generation request.
Training of the current batch overlaps collection of the next batch.
When the gate opens, the new group size is retained until that outstanding batch has arrived, and it is installed before the following generation request is sent.
The number of prompts is updated in the same assignment, which keeps 128 trajectories in the batch.
The actor selects its estimator from the group sizes inside the batch, so the delayed installation prevents one optimizer step from seeing both estimators.

\begin{figure*}[t]
    \centering
    \includegraphics[width=\textwidth,height=0.78\textheight,keepaspectratio]{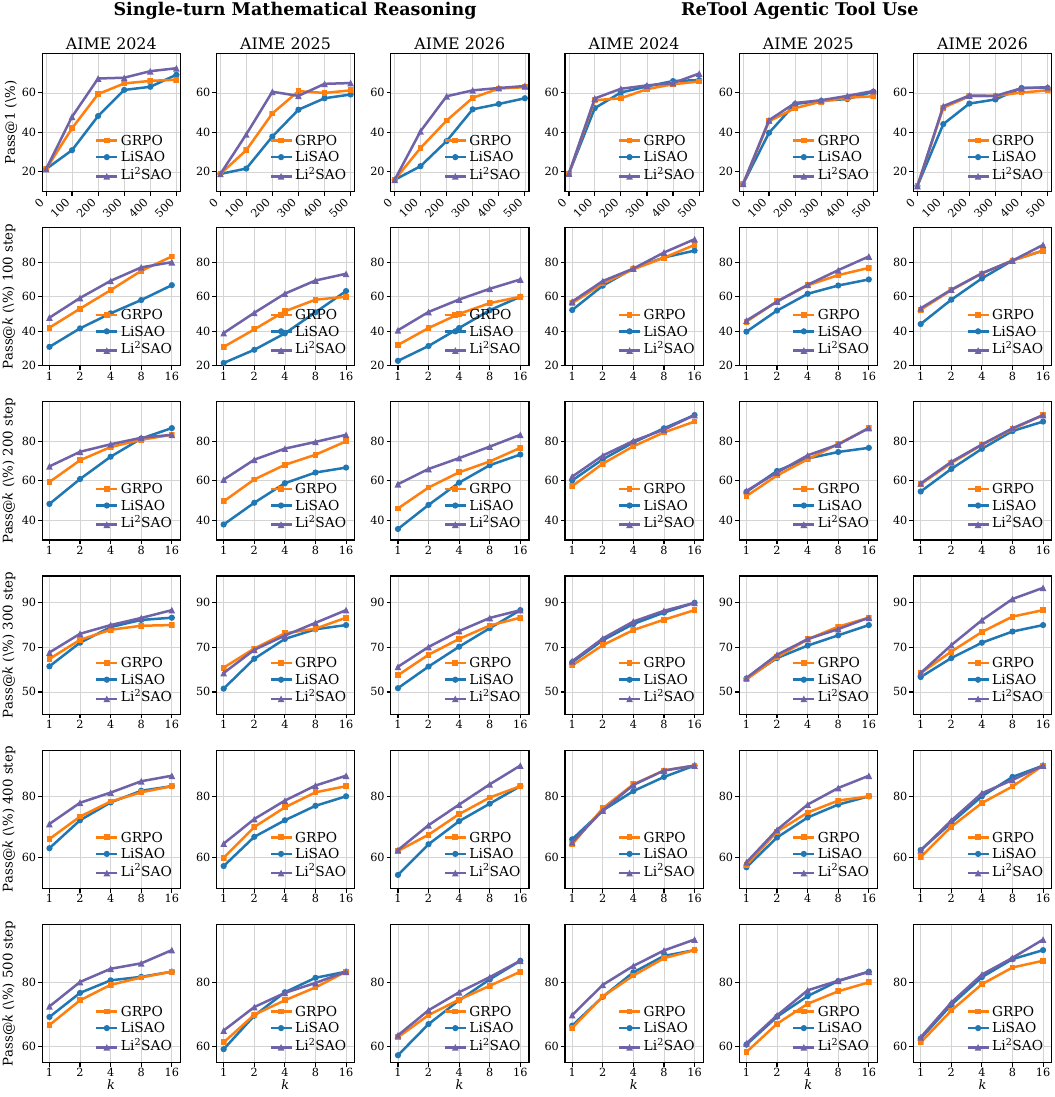}
    \caption{
    \textbf{\liisa against GRPO and \lisao.}
    Results are shown separately for single-turn mathematical reasoning (left) and ReTool agentic tool use (right), and for AIME 2024, 2025, and 2026.
    The top row is pass@1 from the shared initialization to step 500.
    Each lower row is pass@$k$ at the labeled checkpoint.
    }
    \label{fig:li2sao_full}
\end{figure*}

Resuming generation first stops the background rollout worker.
That worker is still sampling at the bootstrap group size after the prefetched batch has been taken, so the groups it still holds follow the old sampling rule.
Finished groups that were not consumed, unfinished generations, and the pending sample buffer are discarded, and the worker is then started again.
A subsequent batch therefore cannot combine leftover bootstrap groups with singleton responses.
A new policy is published only after the outstanding generation has been received, so the weight update does not cut through an active rollout.
Tokens already produced keep their prefix and continue under the updated weights, which is the shared in-flight rule.
Actor updates begin on the first step, so this publication is not delayed by a critic warmup.
The critic runs inside the actor training process with attention frozen, and the value head is trained there together with the MLP blocks.

\section{Additional Experimental Analysis}
\label{app:additional-analysis}

\subsection{Full Comparison}
\label{app:full-comparison}

\begin{figure}[t]
    \centering
    \includegraphics[width=\linewidth]{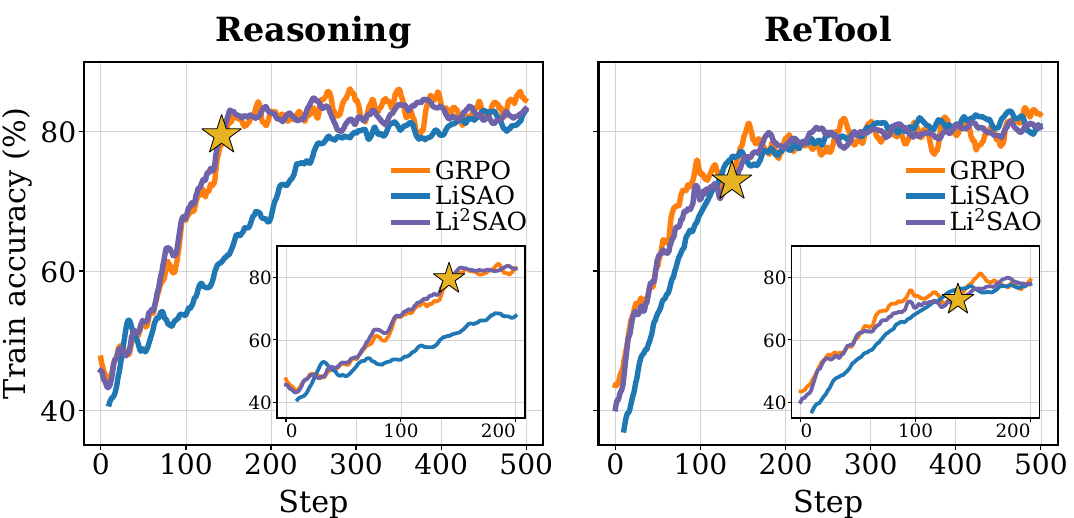}
    \caption{
    \textbf{Training accuracy} of GRPO, \lisao, and \liisa.
    Curves are a two-sided EMA (span 10) of per-step 0/1 accuracy on the training batches.
    The inset shows the first 200 steps.
    Gold stars mark the switch, at step 142 on reasoning and step 137 on ReTool.
    \lisao starts at step 10, after critic warmup.
    }
    \label{fig:train_dynamics}
\end{figure}

Figure~\ref{fig:li2sao_full} places \liisa on the same per-year axes as GRPO and \lisao, supplementing the averages in Section~\ref{sec:exp}.
Single-turn mathematical reasoning occupies the left three columns, and ReTool occupies the right three.
The top row follows pass@1 from the shared initialization through step 500.
Each lower row shows pass@$k$ at one checkpoint, with $k\in\{1,2,4,8,16\}$.

\paragraph{Single-turn reasoning.}
Averaged over AIME 2024, 2025, and 2026, \liisa leads both baselines at every checkpoint and every $k$.
The pass@1 margin is 7.4 percentage points over GRPO and 17.3 over \lisao at step 100, and 10.4 and 21.5 points at step 200.
Group-relative bootstrapping removes the early deficit of \lisao without value pretraining, and the margin at step 200 shows that the gain remains after the switch to single-rollout GAE.
Over GRPO the margin is 1.4 points at step 300 and 3.4 points at step 500.
Over \lisao it is 5.1 points at step 500, and average pass@16 is 86.7\%.
Section~\ref{sec:lisao} traces GRPO's lead on reasoning to a single terminal reward, which already gives a group of solutions its baseline.
The later checkpoints agree with that account: the margin narrows, and \liisa still finishes above both baselines.

\paragraph{ReTool.}
Averaged over the three years, \liisa leads both baselines at every checkpoint and every $k$, with a smaller pass@1 margin.
Over GRPO it is 0.8 percentage points at step 100, 2.6 at step 200, and 2.8 at step 500.
Over \lisao it is 6.8 points at step 100 and 1.4 points at step 500.
Section~\ref{sec:lisao} finds that critic noise appears immediately on tool use and lasts longer than on reasoning.
The step-100 margin over \lisao is that deficit closed by the bootstrap, and the step-200 margin shows the same lead after the switch.
Drawing more samples widens the final gap.
At step 500, average pass@16 is 90.0\%, exceeding GRPO by 4.4 points and \lisao by 2.2 points, and it reaches 93.3\% on AIME 2024 and AIME 2026.
Single-rollout GAE therefore keeps the late advantage of value-based credit assignment on this task, and the early gap is already closed.

\subsection{Training Dynamics}
\label{app:training-dynamics}

\begin{figure}[t]
    \centering
    \includegraphics[width=\linewidth]{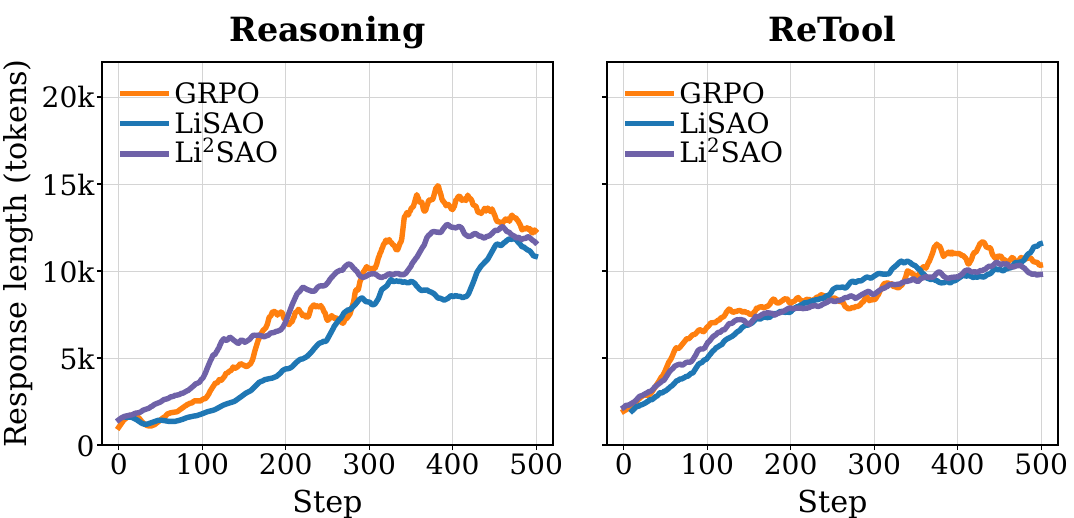}
    \caption{
    \textbf{Mean response length} of GRPO, \lisao, and \liisa.
    Curves are a two-sided EMA (span 10) of per-step mean response length on the training batches.
    \lisao starts at step 10, after critic warmup.
    }
    \label{fig:train_response_len}
\end{figure}

Figure~\ref{fig:train_dynamics} plots training-batch accuracy for the runs in Section~\ref{sec:exp}.
Figure~\ref{fig:train_response_len} plots mean response length, with GRPO and \lisao taken from those runs.
Both series are the two-sided EMA of span 10 given in the captions.
Actor updates of \liisa begin at step 0, together with the critic.
\lisao enters both plots at step 10, after critic warmup.
Through these training dynamics, we analyze the optimization path at a finer grain than the checkpoint evaluations in Section~\ref{sec:exp}, tracing how group-relative bootstrapping removes the early deficit of Section~\ref{sec:lisao} and how the marked switch keeps that repair once the actor moves to single-rollout GAE.
The length curves then record how many tokens each run spends along training.

\paragraph{Single-turn reasoning.}
GRPO and \liisa stay aligned through the bootstrap, while \lisao falls behind after about step 50 and is still near 54\% at step 100, against about 67\% and 68\%.
This is the training-batch form of the early gap in Section~\ref{sec:lisao}.
Explained variance is already positive by then, but one terminal reward leaves little structure inside a solution, so the group baseline remains the useful signal and group-relative advantages keep \liisa on the GRPO path.
The switch at step 142 keeps that gain.
Through step 200, \liisa stays near 83\% with GRPO, while \lisao is still near 68\% and joins the same band only after roughly step 400.
\liisa therefore enters single-rollout GAE from a GRPO-level policy, the route to the higher final pass@$k$ in Table~\ref{tab:main-results}.

\paragraph{ReTool.}
The deficit of \lisao appears at once, as in Figure~\ref{fig:early_train_reward}, because critic noise lasts longer on tool use.
At step 25 it is about 44\%, while GRPO and \liisa are already near 55\%.
Group-relative bootstrapping closes that immediate deficit.
By the switch at step 137, \lisao has caught up to about 76\%, with the other two near 73\%, following the critic recovery in Figure~\ref{fig:critic_dynamics}.
Once the critic tracks the return, skip-observation GAE can separate an early plan, a tool call, and a later mistake.
The three curves then stay within about five percentage points and finish near 81\%.
The handoff preserves the repaired trajectory, and the remaining updates use the single-rollout estimator that Section~\ref{sec:lisao} links to value-based credit assignment on tool use.

\paragraph{Response length.}
Figure~\ref{fig:train_response_len} shows how each estimator spends tokens.
On reasoning, \lisao stays short while the cold critic leaves its updates weak, and it lengthens only as training accuracy recovers.
\liisa, using a group-relative advantage, lengthens earlier than both baselines.
GRPO begins on a similarly short curve and then extends the fastest, becoming the longest trace later in training.
This ordering agrees with Section~\ref{sec:lisao}.
A sequence-level advantage is written on every token and keeps adding length, while one terminal reward gives GAE little structure with which to grow the solution until the critic is useful.
On ReTool the three curves rise together.
\lisao is short during its early deficit and becomes the longest once the critic tracks the return.
\liisa stays with GRPO through the rise and finishes slightly shorter.
Multi-turn traces keep the tool-use turns that skip-observation GAE separates, so the later phase changes credit inside the trajectory while length continues to rise.

\subsection{Ablations}
\label{app:ablations}

\begin{table*}[t]
    \centering
    \footnotesize
    \setlength{\tabcolsep}{1.6pt}
    \renewcommand{\arraystretch}{1.08}
    \caption{
    \textbf{Ablations of \liisa.}
    Each row keeps the group-relative actor and the readiness switch, and changes one critic target.
    The three tables report AIME 2024, 2025, and 2026 separately.
    Pass@$k$ at steps 100, 200, 300, 400, and 500 uses 16 samples per problem at temperature 1.0 and the unbiased estimator of \citet{chen-2021-codex}.
    }
    \label{tab:ablations}
    \parbox{\textwidth}{\centering\textbf{AIME 2024}}\\[0.35em]
    \resizebox{\textwidth}{!}{%
    \begin{tabular}{@{} l @{\hspace{0.7em}} *{5}{c} @{\hspace{0.4em}} *{5}{c} @{\hspace{0.4em}} *{5}{c} @{\hspace{0.4em}} *{5}{c} @{\hspace{0.4em}} *{5}{c} @{}}
        \toprule
        & \multicolumn{5}{c}{\textbf{Step 100}}
        & \multicolumn{5}{c}{\textbf{Step 200}}
        & \multicolumn{5}{c}{\textbf{Step 300}}
        & \multicolumn{5}{c}{\textbf{Step 400}}
        & \multicolumn{5}{c}{\textbf{Step 500}} \\
        \cmidrule(lr){2-6} \cmidrule(lr){7-11} \cmidrule(lr){12-16} \cmidrule(lr){17-21} \cmidrule(lr){22-26}
        \textbf{Variant}\hspace{1.5em}Pass@$k$
        & 1 & 2 & 4 & 8 & 16
        & 1 & 2 & 4 & 8 & 16
        & 1 & 2 & 4 & 8 & 16
        & 1 & 2 & 4 & 8 & 16
        & 1 & 2 & 4 & 8 & 16 \\
        \midrule
        \multicolumn{26}{@{}l@{}}{\cellcolor{bandblue}\rule[-0.42ex]{0pt}{2.35ex}\textit{Single-turn Mathematical Reasoning}} \\
        \liisa
        & 47.9 & 59.3 & 69.2 & 77.0 & 80.0
        & 67.3 & 74.7 & 78.5 & 81.8 & 83.3
        & 67.7 & 76.1 & 80.0 & 83.2 & 86.7
        & 71.0 & 77.9 & 81.2 & 84.9 & 86.7
        & 72.5 & 80.1 & 84.2 & 85.9 & 90.0 \\
        w/o residual target
        & 40.8 & 50.0 & 58.8 & 68.9 & 80.0
        & 57.1 & 66.2 & 71.3 & 73.2 & 73.3
        & 61.0 & 70.5 & 77.6 & 83.0 & 86.7
        & 65.0 & 72.5 & 77.3 & 81.0 & 83.3
        & 68.1 & 74.6 & 77.1 & 78.3 & 80.0 \\
        w/o ranking loss
        & 46.3 & 57.4 & 67.9 & 75.9 & 80.0
        & 60.8 & 70.7 & 77.6 & 81.2 & 83.3
        & 65.0 & 74.0 & 79.5 & 81.6 & 83.3
        & 68.5 & 76.8 & 81.3 & 83.0 & 83.3
        & 66.9 & 76.5 & 80.9 & 82.6 & 83.3 \\
        \midrule
        \multicolumn{26}{@{}l@{}}{\cellcolor{bandblue}\rule[-0.42ex]{0pt}{2.35ex}\textit{ReTool Agentic Tool Use}} \\
        \liisa
        & 57.2 & 69.1 & 76.3 & 85.6 & 93.3
        & 62.1 & 72.8 & 80.2 & 85.9 & 93.3
        & 63.8 & 74.0 & 81.6 & 86.4 & 90.0
        & 65.0 & 75.4 & 83.7 & 88.4 & 90.0
        & 69.8 & 79.2 & 85.1 & 89.9 & 93.3 \\
        w/o residual target
        & 53.8 & 66.1 & 75.3 & 81.2 & 86.7
        & 58.5 & 69.6 & 77.2 & 81.2 & 83.3
        & 62.1 & 73.2 & 80.2 & 84.0 & 86.7
        & 65.0 & 75.0 & 81.3 & 84.7 & 86.7
        & 64.8 & 74.3 & 81.1 & 85.3 & 86.7 \\
        w/o ranking loss
        & 53.1 & 66.2 & 75.4 & 80.4 & 83.3
        & 59.8 & 71.0 & 79.5 & 84.7 & 86.7
        & 63.1 & 73.9 & 81.5 & 85.8 & 90.0
        & 64.4 & 73.7 & 80.0 & 84.6 & 90.0
        & 64.6 & 73.8 & 79.9 & 83.8 & 86.7 \\
        \bottomrule
    \end{tabular}}

    \vspace{1em}
    \parbox{\textwidth}{\centering\textbf{AIME 2025}}\\[0.35em]
    \resizebox{\textwidth}{!}{%
    \begin{tabular}{@{} l @{\hspace{0.7em}} *{5}{c} @{\hspace{0.4em}} *{5}{c} @{\hspace{0.4em}} *{5}{c} @{\hspace{0.4em}} *{5}{c} @{\hspace{0.4em}} *{5}{c} @{}}
        \toprule
        & \multicolumn{5}{c}{\textbf{Step 100}}
        & \multicolumn{5}{c}{\textbf{Step 200}}
        & \multicolumn{5}{c}{\textbf{Step 300}}
        & \multicolumn{5}{c}{\textbf{Step 400}}
        & \multicolumn{5}{c}{\textbf{Step 500}} \\
        \cmidrule(lr){2-6} \cmidrule(lr){7-11} \cmidrule(lr){12-16} \cmidrule(lr){17-21} \cmidrule(lr){22-26}
        \textbf{Variant}\hspace{1.5em}Pass@$k$
        & 1 & 2 & 4 & 8 & 16
        & 1 & 2 & 4 & 8 & 16
        & 1 & 2 & 4 & 8 & 16
        & 1 & 2 & 4 & 8 & 16
        & 1 & 2 & 4 & 8 & 16 \\
        \midrule
        \multicolumn{26}{@{}l@{}}{\cellcolor{bandblue}\rule[-0.42ex]{0pt}{2.35ex}\textit{Single-turn Mathematical Reasoning}} \\
        \liisa
        & 39.0 & 50.7 & 61.8 & 69.4 & 73.3
        & 60.6 & 70.7 & 76.3 & 79.7 & 83.3
        & 58.5 & 68.9 & 75.3 & 81.0 & 86.7
        & 64.6 & 72.6 & 78.6 & 83.5 & 86.7
        & 65.0 & 72.3 & 76.7 & 79.8 & 83.3 \\
        w/o residual target
        & 29.4 & 38.4 & 49.0 & 58.7 & 63.3
        & 50.6 & 62.5 & 71.7 & 77.0 & 80.0
        & 54.0 & 65.5 & 73.0 & 77.2 & 80.0
        & 61.5 & 69.1 & 73.3 & 76.8 & 80.0
        & 61.7 & 70.9 & 76.5 & 80.5 & 83.3 \\
        w/o ranking loss
        & 38.8 & 49.4 & 59.0 & 65.9 & 70.0
        & 50.2 & 61.4 & 68.8 & 73.5 & 76.7
        & 59.2 & 68.4 & 73.2 & 75.6 & 76.7
        & 61.3 & 71.5 & 77.6 & 80.8 & 83.3
        & 60.6 & 69.8 & 75.6 & 79.5 & 83.3 \\
        \midrule
        \multicolumn{26}{@{}l@{}}{\cellcolor{bandblue}\rule[-0.42ex]{0pt}{2.35ex}\textit{ReTool Agentic Tool Use}} \\
        \liisa
        & 46.2 & 57.2 & 66.9 & 75.4 & 83.3
        & 54.9 & 64.2 & 72.9 & 78.4 & 86.7
        & 56.3 & 66.7 & 73.7 & 78.2 & 83.3
        & 58.5 & 69.1 & 77.3 & 82.7 & 86.7
        & 61.0 & 69.7 & 77.5 & 80.5 & 83.3 \\
        w/o residual target
        & 39.6 & 51.0 & 60.1 & 66.3 & 70.0
        & 49.4 & 61.1 & 69.1 & 73.8 & 76.7
        & 55.4 & 66.0 & 72.1 & 75.5 & 76.7
        & 56.7 & 66.6 & 73.2 & 77.9 & 80.0
        & 56.9 & 66.2 & 71.7 & 75.2 & 76.7 \\
        w/o ranking loss
        & 45.2 & 56.6 & 64.9 & 70.2 & 73.3
        & 53.1 & 64.6 & 72.7 & 78.4 & 83.3
        & 54.6 & 65.4 & 72.4 & 77.3 & 80.0
        & 56.7 & 67.8 & 74.6 & 78.1 & 80.0
        & 58.1 & 68.1 & 74.7 & 79.1 & 83.3 \\
        \bottomrule
    \end{tabular}}

    \vspace{1em}
    \parbox{\textwidth}{\centering\textbf{AIME 2026}}\\[0.35em]
    \resizebox{\textwidth}{!}{%
    \begin{tabular}{@{} l @{\hspace{0.7em}} *{5}{c} @{\hspace{0.4em}} *{5}{c} @{\hspace{0.4em}} *{5}{c} @{\hspace{0.4em}} *{5}{c} @{\hspace{0.4em}} *{5}{c} @{}}
        \toprule
        & \multicolumn{5}{c}{\textbf{Step 100}}
        & \multicolumn{5}{c}{\textbf{Step 200}}
        & \multicolumn{5}{c}{\textbf{Step 300}}
        & \multicolumn{5}{c}{\textbf{Step 400}}
        & \multicolumn{5}{c}{\textbf{Step 500}} \\
        \cmidrule(lr){2-6} \cmidrule(lr){7-11} \cmidrule(lr){12-16} \cmidrule(lr){17-21} \cmidrule(lr){22-26}
        \textbf{Variant}\hspace{1.5em}Pass@$k$
        & 1 & 2 & 4 & 8 & 16
        & 1 & 2 & 4 & 8 & 16
        & 1 & 2 & 4 & 8 & 16
        & 1 & 2 & 4 & 8 & 16
        & 1 & 2 & 4 & 8 & 16 \\
        \midrule
        \multicolumn{26}{@{}l@{}}{\cellcolor{bandblue}\rule[-0.42ex]{0pt}{2.35ex}\textit{Single-turn Mathematical Reasoning}} \\
        \liisa
        & 40.6 & 51.2 & 58.4 & 64.6 & 70.0
        & 58.3 & 66.0 & 71.5 & 77.3 & 83.3
        & 61.3 & 70.2 & 77.3 & 83.2 & 86.7
        & 62.5 & 70.6 & 77.3 & 83.9 & 90.0
        & 63.5 & 71.3 & 76.9 & 81.6 & 86.7 \\
        w/o residual target
        & 33.3 & 44.3 & 53.7 & 61.1 & 66.7
        & 50.0 & 58.8 & 65.3 & 71.4 & 76.7
        & 53.1 & 62.5 & 69.6 & 76.0 & 83.3
        & 59.6 & 67.9 & 75.5 & 81.9 & 86.7
        & 60.2 & 68.1 & 75.1 & 79.6 & 80.0 \\
        w/o ranking loss
        & 35.2 & 47.3 & 57.0 & 63.1 & 66.7
        & 48.1 & 57.8 & 65.7 & 71.9 & 76.7
        & 58.1 & 66.9 & 73.3 & 76.4 & 76.7
        & 61.0 & 69.4 & 76.1 & 80.6 & 83.3
        & 60.6 & 67.9 & 73.9 & 79.0 & 83.3 \\
        \midrule
        \multicolumn{26}{@{}l@{}}{\cellcolor{bandblue}\rule[-0.42ex]{0pt}{2.35ex}\textit{ReTool Agentic Tool Use}} \\
        \liisa
        & 53.3 & 64.1 & 73.6 & 80.9 & 90.0
        & 58.7 & 69.6 & 78.4 & 86.6 & 93.3
        & 58.5 & 71.0 & 82.2 & 91.7 & 96.7
        & 62.4 & 72.1 & 81.0 & 85.3 & 90.0
        & 62.9 & 73.7 & 82.4 & 87.6 & 93.3 \\
        w/o residual target
        & 46.3 & 58.8 & 70.5 & 80.9 & 90.0
        & 55.4 & 68.1 & 78.4 & 86.6 & 93.3
        & 57.3 & 68.2 & 76.6 & 82.5 & 86.7
        & 58.3 & 66.9 & 74.6 & 81.7 & 90.0
        & 59.2 & 67.9 & 76.8 & 84.7 & 90.0 \\
        w/o ranking loss
        & 47.9 & 60.4 & 70.3 & 78.9 & 86.7
        & 54.0 & 63.8 & 72.1 & 80.2 & 86.7
        & 58.3 & 66.1 & 74.4 & 82.7 & 90.0
        & 59.4 & 70.0 & 79.1 & 86.3 & 90.0
        & 57.3 & 67.7 & 76.0 & 82.8 & 86.7 \\
        \bottomrule
    \end{tabular}}
\end{table*}

Table~\ref{tab:ablations} isolates the two components of the critic objective while keeping the group relative actor, the readiness switch, the trajectory budget, and the in flight pipeline fixed.
Removing the group relative phase recovers \lisao, while suppressing the switch recovers GRPO, both of which are included in Table~\ref{tab:main-results}.
We therefore focus on the design choices that determine whether the critic can reliably inherit the policy update after bootstrapping.

\paragraph{Critic target.}
\textit{w/o residual target} fits \(V_\phi\) to the absolute return \(\hat G_{i,t}\) during bootstrapping.
Because the divergence mask and ranking loss are defined on mixed groups with centered targets, both are disabled in this variant.
\textit{w/o ranking loss} retains the centered target and divergence mask but removes the pairwise ordering objective.

\paragraph{Residual supervision enables a reliable handoff.}
The complete method consistently outperforms the absolute return variant across benchmark years and interaction regimes.
The gap emerges during bootstrapping and generally persists after the actor switches to critic based advantages.
This pattern indicates that absolute return regression spends critic capacity on prompt dependent difficulty and reward offsets that are irrelevant to the within prompt comparisons used by the initial actor.
Centering the target removes this nuisance variation and the divergence mask excludes groups that provide no comparative signal.
The resulting critic is better aligned with the group relative policy update, which makes its predictions more dependable when it takes over advantage estimation.

\paragraph{Ranking supervision improves discrimination and stability.}
Removing the ranking objective produces a milder but systematic degradation.
The variant can track the full method early in training, yet it more often plateaus or loses ground after the handoff.
Residual regression alone constrains prediction magnitudes but does not directly enforce the ordering between successful and unsuccessful responses.
The ranking objective supplies this missing structure and encourages the critic to preserve the preference relation exposed by mixed outcome groups.
Its benefit across both direct reasoning and tool interaction suggests that the objective improves critic transfer rather than exploiting properties of a particular response format.

\paragraph{The critic objectives are complementary.}
Taken together, the ablations show that the gains do not arise from adding an arbitrary value function to a group based learner.
The residual target aligns the critic with the comparative signal used during bootstrapping, while the ranking objective sharpens that signal into a robust ordering.
Their combination yields the most consistent learning trajectory across tasks and benchmark years, supporting the central premise that a structured critic target is necessary for an effective transition from group relative to critic guided optimization.

\subsection{Training Efficiency}
\label{app:training-efficiency}

Table~\ref{tab:train-efficiency} reports the wall-clock time of 500 training steps on 8 A800 GPUs in Appendix~\ref{app:hardware}.
All rows use the in-flight pipeline of Section~\ref{sec:prelim_inflight}, and both device splits draw on this same pool of 8 GPUs.
GRPO keeps eight responses per prompt at every step, \lisao draws one response per prompt, and \liisa starts from the GRPO group size before switching once to one response per prompt.

\begin{table}[t]
    \centering
    \small
    \setlength{\tabcolsep}{3.2pt}
    \renewcommand{\arraystretch}{1.12}
    \caption{
    \textbf{Wall-clock time} of 500 training steps for single-turn mathematical reasoning and ReTool agentic tool use.
    Train\,+\,Rollout is the GPU split between the trainer and the rollout workers.
    In \lisao and \liisa, the actor and the critic share the trainer GPUs.
    Prompts\,$\times$\,Samples is the number of prompts in each step times the responses drawn for each prompt.
    }
    \label{tab:train-efficiency}
    \begin{tabular*}{\columnwidth}{@{\extracolsep{\fill}}lccc@{}}
        \toprule
        \textbf{Method} & \textbf{Train\,+\,Rollout} & \textbf{Prompts\,$\times$\,Samples} & \textbf{Hours} \\
        \midrule
        \multicolumn{4}{@{}l@{}}{\cellcolor{bandblue}\rule[-0.42ex]{0pt}{2.35ex}\textit{Single-turn Mathematical Reasoning}} \\
        GRPO & 4\,+\,4 & 16\,$\times$\,8 & 51.5 \\
        GRPO & 2\,+\,6 & 16\,$\times$\,8 & 36.4 \\
        \lisao & 4\,+\,4 & 128\,$\times$\,1 & 31.8 \\
        \liisa & 4\,+\,4 & 16\,$\times$\,8\,$\to$\,128\,$\times$\,1 & 28.9 \\
        \midrule
        \multicolumn{4}{@{}l@{}}{\cellcolor{bandblue}\rule[-0.42ex]{0pt}{2.35ex}\textit{ReTool Agentic Tool Use}} \\
        GRPO & 4\,+\,4 & 16\,$\times$\,8 & 48.7 \\
        GRPO & 2\,+\,6 & 16\,$\times$\,8 & 39.6 \\
        \lisao & 4\,+\,4 & 128\,$\times$\,1 & 35.5 \\
        \liisa & 4\,+\,4 & 16\,$\times$\,8\,$\to$\,128\,$\times$\,1 & 33.1 \\
        \bottomrule
    \end{tabular*}
\end{table}

\paragraph{Extra rollout GPUs shorten GRPO.}
The two GRPO rows draw the same 16\,$\times$\,8 responses, so the difference is the device split.
Moving 2 GPUs from the trainer to the rollout workers reduces reasoning from 51.5 hours to 36.4 hours (29.3\%) and ReTool from 48.7 hours to 39.6 hours (18.7\%).
A group-relative step forms its advantage only after eight responses to the same prompt are available.
With long-tailed response lengths, group completion is limited by rollout throughput, and the 2 additional rollout GPUs increase that throughput.

\paragraph{\lisao and \liisa have comparable training costs.}
With 4 trainer GPUs and 4 rollout GPUs, \lisao completes training in 31.8 hours on reasoning and 35.5 hours on ReTool, whereas \liisa requires 28.9 and 33.1 hours, respectively.
Although \liisa begins with group-relative updates at 16\,$\times$\,8, its initial group-based phase does not increase the total training time relative to the 128\,$\times$\,1 configuration used by \lisao from the outset.
This result follows from the evolution of response length shown in Figure~\ref{fig:train_response_len}.
The group-relative bootstrapping phase occurs while the mean response length remains low.
By the time responses become longer, \liisa has switched to 128\,$\times$\,1, so each prompt requires only one of the expensive generations.

\paragraph{Critic training preserves the efficiency advantage over GRPO.}
\lisao and \liisa train the critic jointly with the actor on the same 4 trainer GPUs, whereas GRPO updates only the actor.
Under the 4\,+\,4 split, \lisao reduces the training time relative to GRPO by 19.7 hours on reasoning and 13.2 hours on ReTool.
The corresponding reductions for \liisa are 22.6 and 15.6 hours, which amount to 43.9\% and 32.0\%, respectively.
This advantage persists when GRPO is allocated the more rollout-intensive 2\,+\,6 split.
Even under this configuration, \lisao remains faster by 4.6 hours on reasoning and 4.1 hours on ReTool, while \liisa remains faster by 7.5 and 6.5 hours.
All methods collect 128 trajectories per step, but GRPO can form its group-relative advantage only after all eight responses to a given prompt become available.
As Figure~\ref{fig:train_response_len} indicates, the resulting group-completion delay increases as responses become longer.
By contrast, each completed trajectory in the single-response setting can enter the training batch immediately, which is suited to asynchronous execution.
The scheduling gains therefore outweigh the additional cost of critic optimization.

\subsection{Staleness Dynamics}
\label{app:staleness-dynamics}

\begin{figure}[t]
    \centering
    \includegraphics[width=\linewidth]{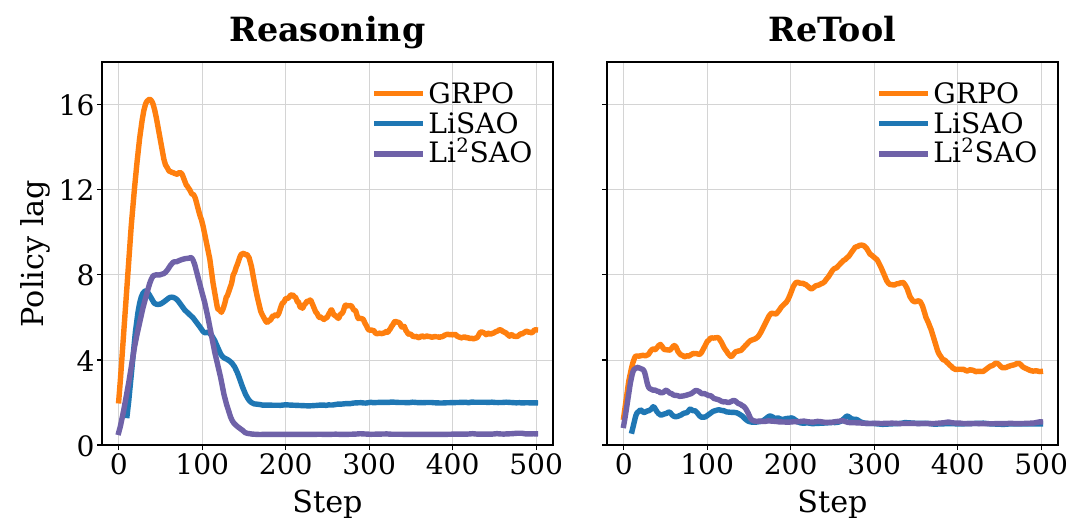}
    \caption{
    \textbf{Mean policy lag} of GRPO, \lisao, and \liisa.
    Curves are a two-sided EMA (span 10) of the per-step mean policy lag.
    GRPO is the training-batch mean.
    \lisao and \liisa are the rollout mean, one policy version lower on runs that log both.
    \lisao starts at step 10, after critic warmup.
    }
    \label{fig:policy_lag}
\end{figure}

\paragraph{Policy lag.}
Figure~\ref{fig:policy_lag} shows how many policy versions the samples of each step trail the current policy.
Across both tasks, GRPO exhibits the largest lag, \lisao remains substantially fresher, and \liisa moves from the group-based regime toward \lisao or below it after the switch.
A group-relative update cannot use an early response until the other seven responses to the same prompt finish; while it waits, the trainer continues to advance, turning response-length variation within a group into policy-version lag.
Single-response collection removes this slowest-member barrier, so a completed trajectory can enter the next available batch without waiting for matched responses.
The sharp within-run decline of \liisa after switching supports this mechanism and shows that single-response sampling, rather than the critic alone, is what reduces policy staleness.

\begin{figure}[t]
    \centering
    \includegraphics[width=\linewidth]{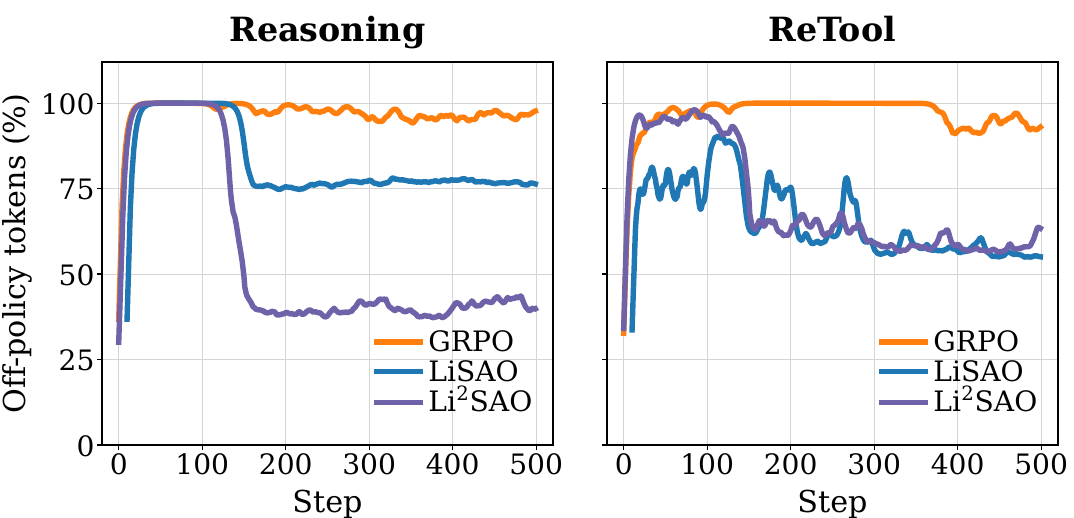}
    \caption{
    \textbf{Off-policy token ratio} of GRPO, \lisao, and \liisa.
    Curves are a two-sided EMA (span 10) of the fraction of response tokens generated more than one policy version behind the trainer.
    A one-step lag is counted as on-policy.
    Every curve is the training-batch ratio.
    \lisao starts at step 10, after critic warmup.
    }
    \label{fig:offpolicy_token_ratio}
\end{figure}

\paragraph{Off-policy tokens.}
Figure~\ref{fig:offpolicy_token_ratio} applies a token-level cut to the same staleness.
A response token is off-policy only when the policy that generated it is more than one version behind the trainer.
All methods are highly off-policy early in training, but GRPO remains near saturation whereas \lisao becomes markedly less stale.
\liisa again behaves like a group-based method during bootstrapping and drops sharply after switching, reaching a level comparable to or lower than \lisao.
This pattern confirms that removing the group-completion barrier not only shortens waiting but also reduces the fraction of stale tokens actually used for learning.
Together, the two figures show that the single-response phases of \lisao and \liisa substantially mitigate off-policy data staleness relative to GRPO.
This conclusion is specifically about policy-version freshness under the stated threshold; it does not by itself imply that importance weighting eliminates all off-policy estimation bias.

\section{Coding Agentic RL}
\label{app:coding-agentic-rl}
To assess the effectiveness of \liisa on more complex agentic tasks and with larger models, we further evaluate the optimization methods on long-horizon software engineering tasks.

\subsection{Infrastructure and Agent Environment}
Our implementation uses OpenHands~\citep{wang-2025-openhands}\footnote{\url{https://github.com/OpenHands/OpenHands}} as the agent scaffold and a modified version of \textsc{slime}\footnote{\url{https://github.com/THUDM/slime}} as the asynchronous RL backend.
Each rollout runs in an isolated Docker environment initialized from the instance-specific repository and base commit.
OpenHands interacts with the repository through its shell and file-editing tools, while an OpenAI-compatible adapter routes every model call to the SGLang rollout engines.
Each trajectory is limited to 300 interaction turns and \(3600\) seconds, with a \(128\)K context window and OpenHands context compression enabled so that long multi-turn traces remain within this budget.
The adapter records the exact generated token IDs, loss masks, and behavior log probabilities for every assistant turn, avoiding a separate re-tokenization step when the multi-turn trajectory is converted into training samples.
This preserves alignment between sampled actions and the tokens used for policy optimization across turns.

The principal systems modification is \emph{agent-step-level in-flight weight synchronization}.
In the original \textsc{slime}, a weight update aborts an in-flight generation request and truncates it before any further tokens are sampled under the new weights.
Our modified \textsc{slime} can instead refresh the rollout-engine weights at agent-step boundaries while a long-running OpenHands trajectory is still in flight.
Thus, later agent steps need not wait for the entire trajectory to finish before using a newer policy version.
The behavior log probabilities associated with each generated segment are retained for importance-ratio computation.

All coding experiments use eight nodes with eight NVIDIA A800 GPUs per node, for a total of 64 GPUs.
By default, four nodes are allocated to training and the other four to rollout.
The same accelerator allocation and the same asynchronous rollout infrastructure are used for every RL method.
Table~\ref{tab:coding-setup} summarizes the shared configuration.
To scale environment execution, we implement a distributed rollout manager that deploys one sandbox worker on each of the eight nodes.
The manager dispatches OpenHands trajectories with per node concurrency control and backpressure, while each worker launches its sandbox and reward evaluation containers through the local Docker daemon.
This design avoids a centralized container bottleneck and enables concurrent environment execution across all eight nodes.

\begin{table}[t]
    \centering
    \small
    \caption{Shared configuration for coding agentic RL.}
    \label{tab:coding-setup}
    \begin{tabular*}{0.95\columnwidth}{@{\hspace{\tabcolsep}\extracolsep{\fill}}ll@{\hspace{\tabcolsep}}}
        \toprule
        \textbf{Setting} & \textbf{Value} \\
        \midrule
        Agent scaffold & OpenHands \\
        RL backend & modified \textsc{slime} \\
        Weight synchronization & agent-step-level, in flight \\
        Sandbox & instance-specific Docker \\
        Compute & $8$ nodes $\times$ $8$ A800 GPUs \\
        \midrule
        Context length & \(128\)K \\
        Context compression & enabled \\
        Max interaction turns & 300 \\
        Instance timeout & \(3600\)\,s \\
        Sandbox concurrency & 256 \\
        Max policy lag & 8 \\
        \midrule
        Training data & SWE-Gym + SWE-smith \\
        Reward & binary test outcome \\
        Training steps & 200 \\
        Global batch size & 256 trajectories \\
        Evaluation & SWE-bench Verified \\
        \bottomrule
    \end{tabular*}
\end{table}

\subsection{Training Data and Reward}

We evaluate on SWE-bench Verified~\citep{jimenez-2024-swebench}\footnote{\url{https://github.com/SWE-bench/SWE-bench}} for all coding experiments.
Training uses a mixture of SWE-Gym~\citep{pan-2025-swegym}\footnote{\url{https://github.com/SWE-Gym/SWE-Gym}} and SWE-smith~\citep{yang-2025-swesmith}\footnote{\url{https://github.com/SWE-bench/SWE-smith}}.
Each training instance provides a problem statement, a repository snapshot, a base commit, an environment, and tests that determine whether the issue has been resolved.
The agent produces a repository patch through multi-turn interaction with OpenHands.

We use the sparse binary reward implemented by the coding pipeline.
Scoring is performed in a fresh container: the repository is reset to the prescribed base commit, the model patch is applied, protected test files are restored, and the instance tests are then executed.
The reward is $1$ only if all fail-to-pass tests succeed and all pass-to-pass tests remain successful, and $0$ otherwise.
Rollouts that fail before producing a valid interaction trace, for example because the sandbox cannot be created or the model endpoint is unreachable, are marked as aborted and excluded from optimization rather than treated as negative solutions.

\subsection{Models and Baselines}

We train Qwen3-30B-A3B-Instruct~\citep{yang-2025-qwen3}\footnote{\url{https://huggingface.co/Qwen/Qwen3-30B-A3B-Instruct-2507}} and Qwen3-Coder-30B-A3B~\citep{yang-2025-qwen3}\footnote{\url{https://huggingface.co/Qwen/Qwen3-Coder-30B-A3B-Instruct}}.
For Qwen3-30B-A3B-Instruct, we explicitly disable its thinking mode during both training and evaluation.
Qwen3-Coder-30B-A3B does not expose a separate thinking mode, so we use its native non-thinking configuration.
This keeps the interaction protocol comparable and prevents an additional hidden-reasoning budget from confounding the algorithmic comparison.

For each model, we compare six settings.
\textsc{Base} evaluates the unmodified instruction checkpoint without RL.
GRPO uses prompt-level group-relative advantages~\citep{shao-2024-deepseekmath}.
DAPO adds dynamic sampling and asymmetric clipping~\citep{yu-2025-dapo}.
GSPO applies sequence-level importance weighting and clipping~\citep{zheng-2025-gspo}.
\lisao uses a learned critic and single-rollout GAE throughout training.
\liisa first performs group-relative actor-critic bootstrapping and then switches to single-rollout GAE according to the procedure in Section~\ref{sec:method}, with the switch window set to \((T_{\min},T_{\max})=(30,60)\).
All RL methods follow the shared settings in Table~\ref{tab:coding-setup} and are trained for 200 steps with a global batch of 256 trajectories: group-based methods use \(32\times 8\), \lisao uses \(256\times 1\), and \liisa starts from \(32\times 8\) and switches to \(256\times 1\).
Except for their algorithm-specific advantage estimators, clipping rules, and critic updates, all methods use the same data, agent environment, reward, and total accelerator allocation.
The mathematical experiments are designed to compare algorithms whose advantage estimators differ in kind, so they omit DAPO and GSPO, which remain GRPO variants and still rely on group-relative advantages.
We include them here because SWE-bench groups often receive all-$1$ or all-$0$ rewards, so DAPO's dynamic sampling can preserve a nonzero advantage at several times the rollout cost, and because GSPO's sequence-level weighting targets the mixture-of-experts models trained in this section.

\subsection{Evaluation Protocol and Results}

We evaluate every checkpoint on SWE-bench Verified with the same OpenHands scaffold and Docker execution protocol used during training.
For each instance, the model must inspect the repository, implement a patch, and validate the change through tool interaction.
We report the official resolved rate, namely the percentage of the 500 verified instances whose fail-to-pass tests are fixed without regressions in pass-to-pass tests.
Table~\ref{tab:coding-results} provides the comparison, including the two critic-target ablations of \liisa from Appendix~\ref{app:ablations}.

\begin{table}[t]
    \centering
    \small
    \renewcommand{\arraystretch}{1.12}
    \caption{
    \textbf{SWE-bench Verified resolved rate} (\%) after coding agentic RL on SWE-Gym.
    }
    \label{tab:coding-results}
    \begin{tabularx}{0.95\columnwidth}{@{\hspace{\tabcolsep}}X c@{\hspace{\tabcolsep}}}
        \toprule
        \textbf{Method} & \textbf{Resolved Rate (\%)} \\
        \midrule
        \multicolumn{2}{@{}l@{}}{\cellcolor{bandblue}\rule[-0.42ex]{0pt}{2.35ex}\textit{Qwen3-30B-A3B-Instruct (thinking off)}} \\
        Base & 21.6 \\
        GRPO & 24.6\\
        DAPO & 24.2\\
        GSPO & 25.8\\
        \lisao & 25.4\\
        \midrule
        \rowcolor{rowgray} \liisa & \textbf{27.2}\\
        \quad w/o residual target & 26.2  \\
        \quad w/o ranking loss & 26.6 \\
        \midrule
        \multicolumn{2}{@{}l@{}}{\cellcolor{bandblue}\rule[-0.42ex]{0pt}{2.35ex}\textit{Qwen3-Coder-30B-A3B (non-thinking)}} \\
        Base & 49.2 \\
        GRPO & 52.8 \\
        DAPO & 53.0\\
        GSPO & 53.6 \\
        \lisao & 53.2 \\
        \midrule
        \rowcolor{rowgray} \liisa & \textbf{54.2}\\
        \quad w/o residual target & 53.4 \\
        \quad w/o ranking loss & 53.6 \\
        \bottomrule
    \end{tabularx}
\end{table}

\paragraph{\liisa reaches the highest resolved rate on both models.}
GRPO, DAPO, and GSPO improve both checkpoints and stay close to one another.
DAPO remains comparable to GRPO, and GSPO is the strongest of the three while staying in the same band, so sequence-level advantages raise the policy and then share a ceiling on these long tool traces.
\lisao, trained from a cold critic without value pretraining, already joins that band and finishes ahead of GRPO.
\liisa is best on both the general instruction model and the code model.
Its margin over \lisao comes from group-relative actor-critic bootstrapping, and its margin over the group-relative methods comes from the switch to single-rollout GAE.
The lead narrows on the code checkpoint, while the ranking stays the same.

\paragraph{Both critic targets are required for this handoff.}
Removing the residual target hurts more than removing the ranking loss on both models.
Centering keeps the critic on which patch succeeds, and ranking keeps successful traces above failed ones.
On the instruction model, both ablations still finish above \lisao and every group-relative method.
On the code model, both ablations fall back to the band of GSPO and \lisao, and only the full objective remains ahead.
Group-relative bootstrapping followed by one switch to single-rollout GAE raises the resolved rate on long-horizon software engineering without value pretraining.

\end{document}